%% file: main.tex
\documentclass[12pt]{article}

\usepackage[top=0.95in,bottom=0.95in,left=0.95in,right=0.95in]{geometry}
\usepackage[utf8]{inputenc} 
\usepackage{lmodern}
\usepackage[T1]{fontenc}    
\usepackage{hyperref}      
\usepackage{url}            
\usepackage{booktabs}       
\usepackage{amsfonts}       
\usepackage{nicefrac}       
\usepackage{microtype}  

\input{macros}

\begin{document}

\title{Fair Performance Metric Elicitation}

\author{Gaurush Hiranandani \\
UIUC\\
gaurush2@illinois.edu
\and
Harikrishna Narasimhan \\
Google Research\\
hnarasimhan@google.com
\and
Oluwasanmi Koyejo\\
UIUC \& Google Research Accra\\
sanmi@illinois.edu
}

\date{\today}

\flushbottom
\maketitle

\input{abstract}

\input{introduction}

\input{background}

\input{confusion}

\input{elicitation}

\input{guarantees}

\input{experiments}

\input{relatedwork}

\input{discussion}

\input{conclusion}

\bibliography{references}
\bibliographystyle{abbrv}

\clearpage

\input{supplement}

\end{document}

%% file: macros.tex
\usepackage{graphicx}
\usepackage{subfigure}
\usepackage{caption}
\usepackage{diagbox}

\usepackage{amsmath}
\usepackage{dsfont}
\usepackage{amssymb,amsthm,commath}
\usepackage{bm,xfrac,xcolor}
\usepackage{nccmath}
\usepackage[toc,page,header]{appendix}

\usepackage{algorithm}
\usepackage[noend]{algorithmic}
\usepackage{bbm}
\usepackage{enumitem}
\usepackage{gensymb}
\usepackage{refcount}
\usepackage{mathtools}
\usepackage{stackengine}
\usepackage{multirow}
\usepackage{makecell}

\usepackage{comment}
\usepackage{mathrsfs}


\newcommand{\INDSTATE}[1][1]{\STATE\hspace{-0.75em}{\algorithmicindent}}

\def\myfnt{\ifx\protect\@typeset@protect\expandafter\footnote\else\expandafter\@gobble\fi}
\makeatother

\makeatletter
\def\BState{\State\hskip-\ALG@thistlm}
\makeatother





\usepackage{placeins,afterpage}

\makeatletter
\newcommand*\bigcdot{\mathpalette\bigcdot@{.5}}
\newcommand*\bigcdot@[2]{\mathbin{\vcenter{\hbox{\scalebox{#2}{$\m@th#1\bullet$}}}}}
\makeatother

\usepackage{tikz}
\tikzset{
  c/.style={every coordinate/.try}
}

\newcommand\numberthis{\addtocounter{equation}{1}\tag{\theequation}}

\mathchardef\mhyphen="2D

\newcommand{\offdiag}{\mathit{off\mhyphen diag}}

\newcommand{\tupr}{\rmbf^{1:m}}

\newcommand{\Hcal}{{\cal H}}

\newcommand{\Lcal}{{\cal L}}
\newcommand{\Mcal}{{\cal M}}

\newcommand{\Rcal}{{\cal R}}
\newcommand{\Scal}{{\cal S}}

\newcommand{\Xcal}{{\cal X}}

\newcommand{\Zcal}{{\cal Z}}

\newcommand{\Pmbb}{\mathbb{P}}

\newcommand{\Rmbb}{\mathbb{R}}

\newcommand{\Zmbb}{\mathbb{Z}}

\newcommand{\thetambf}{\bm{\theta}}

\newcommand{\ambfhat}{\hat{\mathbf{a}}}

\newcommand{\bmbfhat}{\hat{\mathbf{b}}}
\newcommand{\Bmbfhat}{\hat{\mathbf{B}}}
\newcommand{\lambdahat}{\hat{\lambda}}

\newcommand{\ambfbar}{{\mathbf{a}}}
\newcommand{\zmbfbar}{{\mathbf{z}}}

\newcommand{\bmbfbar}{{\mathbf{b}}}
\newcommand{\Bmbfbar}{{\mathbf{B}}}
\newcommand{\lambdabar}{{\lambda}}

\newcommand{\Ambf}{\mathbf{A}}
\newcommand{\ambf}{\mathbf{a}}
\newcommand{\Bmbf}{\mathbf{B}}
\newcommand{\bmbf}{\mathbf{b}}

\newcommand{\dmbf}{\mathbf{d}}

\newcommand{\embf}{\mathbf{e}}

\newcommand{\fmbf}{\mathbf{f}}

\newcommand{\ombf}{\mathbf{o}}

\newcommand{\Rmbf}{\mathbf{R}}
\newcommand{\rmbf}{\mathbf{r}}

\newcommand{\smbf}{\mathbf{s}}

\newcommand{\tmbf}{\mathbf{t}}

\newcommand{\wmbf}{\mathbf{w}}

\newcommand{\xmbf}{\mathbf{x}}

\newcommand{\zmbf}{\mathbf{z}}

\newcommand{\oline}[1]{\mkern 1.5mu\overline{\mkern-1.5mu#1}}

\newcommand{\abar}{\oline{a}}

\newcommand{\bbar}{\oline{b}}

\renewcommand{\hbar}{\oline{h}}

\newcommand{\zbar}{\oline{z}}

\newcommand\inner[2]{\langle #1, #2 \rangle}

\newcommand{\1}{\mathbbm{1}}

\newcommand{\hphi}{\hat{\phi}}
\newcommand{\hvarphi}{\hat{\varphi}}

\newcommand{\hPsi}{\hat{\Psi}}

\newcommand{\bphi}{{\phi}}
\newcommand{\bvarphi}{{\varphi}}

\newcommand{\bPsi}{{\Psi}}

\newcommand{\scl}[1]{\scalebox{0.9}{#1}}

\newcommand{\taumbf}{\bm{\tau}}

\newcommand{\gammambf}{\bm{\gamma}}

\newcommand{\etambfbar}{{\bm{\eta}}}

\newcommand{\etabar}{{\eta}}

\DeclareMathOperator*{\argmax}{argmax}

\newcommand{\baligned}     {\begin{aligned}}
	\newcommand{\ealigned}     {\end{aligned}}
\newcommand{\barray}       {\begin{array}}
	\newcommand{\earray}       {\end{array}}
\newcommand{\bbmatrix}     {\begin{bmatrix}}
	\newcommand{\ebmatrix}     {\end{bmatrix}}
\newcommand{\bcases}       {\begin{cases}}
	\newcommand{\ecases}       {\end{cases}}
\newcommand{\bcenter}      {\begin{center}}
	\newcommand{\ecenter}      {\end{center}}
\newcommand{\bcolumn}      {\begin{column}}
	\newcommand{\ecolumn}      {\end{column}}
\newcommand{\bcolumns}     {\begin{columns}}
	\newcommand{\ecolumns}     {\end{columns}}
\newcommand{\benumerate}   {\begin{enumerate}}
	\newcommand{\eenumerate}   {\end{enumerate}}
\newcommand{\bequation}    {\begin{equation}}
	\newcommand{\eequation}    {\end{equation}}
\newcommand{\bequationn}   {\begin{equation*}}
	\newcommand{\eequationn}   {\end{equation*}}
\newcommand{\bfigure}      {\begin{figure}}
	\newcommand{\efigure}      {\end{figure}}
\newcommand{\bflushright}  {\begin{flushright}}
	\newcommand{\eflushright}  {\end{flushright}}
\newcommand{\bitemize}     {\begin{itemize}}
	\newcommand{\eitemize}     {\end{itemize}}
\newcommand{\bpmatrix}     {\begin{pmatrix}}
	\newcommand{\epmatrix}     {\end{pmatrix}}
\newcommand{\bsubequations}{\begin{subequations}}
	\newcommand{\esubequations}{\end{subequations}}
\newcommand{\btable}       {\begin{table}}
	\newcommand{\etable}       {\end{table}}
\newcommand{\btabular}     {\begin{tabular}}
	\newcommand{\etabular}     {\end{tabular}}
\newcommand{\bvmatrix}     {\begin{vmatrix}}
	\newcommand{\evmatrix}     {\end{vmatrix}}

\newcommand{\bequali}      {\bsubequations\begin{align}}
	\newcommand{\eequali}      {\end{align}\esubequations}

\newtheorem{assumption}{Assumption}
\newtheorem{prop}{Proposition}
\newtheorem{example}{Example}
\newtheorem{remark}{Remark}
\newtheorem{theorem}{Theorem}
\newtheorem{corollary}{Corollary}
\newtheorem{definition}{Definition}
\newtheorem{lemma}{Lemma}

\newcommand{\balgorithm}  {\begin{algorithm}}
	\newcommand{\ealgorithm}  {\end{algorithm}}
\newcommand{\balgorithmic}{\begin{algorithmic}}
	\newcommand{\ealgorithmic}{\end{algorithmic}}
\newcommand{\bassumption} {\begin{assumption}}
	\newcommand{\eassumption} {\end{assumption}}
\newcommand{\bcorollary}  {\begin{corollary}}
	\newcommand{\ecorollary}  {\end{corollary}}
\newcommand{\bdefinition} {\begin{definition}}
	\newcommand{\edefinition} {\end{definition}}
\newcommand{\bexample}    {\begin{example}}
	\newcommand{\eexample}    {\end{example}}
\newcommand{\bproposition}    {\begin{prop}}
	\newcommand{\eproposition}    {\end{prop}}
\newcommand{\blemma}      {\begin{lemma}}
	\newcommand{\elemma}      {\end{lemma}}
\newcommand{\bproblem}    {\begin{problem}}
	\newcommand{\eproblem}    {\end{problem}}
\newcommand{\bproof}      {\begin{proof}}
	\newcommand{\eproof}      {\end{proof}}
\newcommand{\bremark}     {\begin{remark}}
	\newcommand{\eremark}     {\end{remark}}
\newcommand{\btheorem}    {\begin{theorem}}
	\newcommand{\etheorem}    {\end{theorem}}
	
\newcommand{\prodRcal}{\Rcal^{1:m}}

\usepackage{chngcntr}
\counterwithout{equation}{section}
\counterwithout{table}{section}
\counterwithout{figure}{section}
\counterwithout{algorithm}{section}
\usepackage[customcolors]{hf-tikz}
\usepackage{adjustbox}
\hypersetup{colorlinks,urlcolor=blue, citecolor = blue}
\usepackage[normalem]{ulem}
\usepackage{cancel}
\usetikzlibrary{calc,shapes.geometric}

%% file: abstract.tex
\begin{abstract}
\emph{What is a fair performance metric?} We consider the choice of fairness metrics through the lens of metric elicitation -- a principled framework for selecting performance metrics that best reflect implicit preferences. The use of metric elicitation enables a practitioner to tune the performance and fairness metrics to the task, context, and population at hand. Specifically, we  propose a novel strategy to elicit group-fair performance metrics for multiclass classification problems with multiple sensitive groups that also includes selecting the trade-off between predictive performance and fairness violation. The proposed elicitation strategy requires only relative preference feedback and is robust to both finite sample and feedback noise.
\end{abstract}

%% file: introduction.tex
\section{Introduction}
\label{sec:introduction}

Machine learning models are increasingly employed for critical decision-making tasks such as hiring and sentencing~\cite{singla2015learning, angwin2016machine, corbett2017algorithmic, friedler2019comparative, lahoti2019ifair}. Yet, it is increasingly evident that automated decision-making is susceptible to bias, whereby decisions made by the algorithm are unfair to certain subgroups~\cite{barocas2016big,angwin2016machine, chouldechova2017fair, berk2018fairness, lahoti2019ifair}. To this end, a wide variety of group fairness metrics have been proposed – all to reduce discrimination and bias from automated decision-making~\cite{kamishima2012fairness, dwork2012fairness, hardt2016equality, kleinberg2017inherent, woodworth2017learning, menon2018cost}. However, a dearth of formal principles for selecting the  most appropriate metric has highlighted the confusion of experts, practitioners, and end users in deciding which group fairness metric to employ~\cite{zhang2020joint}. This is further exacerbated by the observation that common metrics often lead to contradictory outcomes~\cite{kleinberg2017inherent}. 

\begin{figure}[t]
\begin{minipage}[t]{\textwidth}
    \begin{minipage}[t]{0.50\textwidth}
    \vspace{5pt}
    \centering
    \includegraphics[scale=0.33]{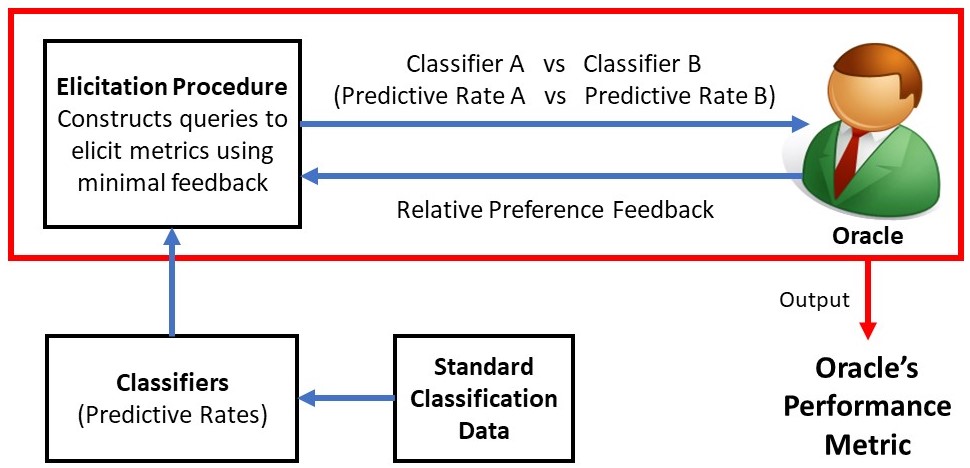}
    \vskip 0.05cm
    \captionof{figure}{Framework of Metric Elicitation~\cite{hiranandani2018eliciting}.}
    \label{fig:ME}
  \end{minipage}\hfill
  \begin{minipage}[t]{0.475\textwidth}
  \vspace{0pt}
    \centering
    \centering
	\begin{tikzpicture}[scale = 1.1]
    \begin{scope}[shift={(4.3,0)},scale = 0.5]\scriptsize
    
    \def\r{0.12};
    
    \coordinate (a) at (-0.2,1);
    \coordinate (b) at (0.8, 2.75);
    \coordinate (c) at (7, 4);
    \coordinate (d) at (6.5, 2);
    \coordinate (e) at (2.5, -0.75);
    \coordinate (f) at (-0.1, -0.5);
    
    \coordinate (Cent) at (3,1.75);
    \coordinate (Centcent) at (2.85,1.90);
    \coordinate (CentR) at (3.6,2.35);
    \coordinate (CentL) at (2.4,1.15);
    
    \coordinate (Space1) at (0.2,0.2);
    \coordinate (Space2) at (-0.3,1.3);
    \coordinate (Spacem) at (-0.1,2.5);
    
    \coordinate (Sphere) at (5,2);
    \coordinate (Sphereplus) at (2.6,2.55);
    \coordinate (Sphereminus) at (3.4,0.95);
    
    \coordinate (r) at (3.75,2);
    
    \coordinate (u11) at (-0.25, -0.25);
    \coordinate (uextra121) at (1.25, 3.75);
    \coordinate (u21) at (4, 4.5);
    \coordinate (uextra2k1) at (6, 1);
    \coordinate (uk1) at (3.5, -0.5);
    
    \coordinate (u12) at (0.3, -0.70);
    \coordinate (uextra122) at (0.65, 2.65);
    \coordinate (u22) at (3.4, 4.2);
    \coordinate (uextra2k2) at (5.15, 1.6);
    \coordinate (uk2) at (3.75, 0.1);
    
    \coordinate (u13) at (-0.30, 0.10);
    \coordinate (uextra123) at (1.25, 3.25);
    \coordinate (u23) at (4, 4.5);
    \coordinate (uextra2k3) at (5.75, 1);
    \coordinate (uk3) at (3.75, -0.5);
    
    \coordinate (u14) at (-0.30, 0.10);
    \coordinate (uextra124) at (1.25, 3.25);
    \coordinate (u24) at (4, 4.5);
    \coordinate (uextra2k4) at (5.75, 1);
    \coordinate (uk4) at (3.75, -0.5);

    \fill[color=black] 
            (Cent) circle (0.08)
            
            (u11) circle (0.08)
            (u21) circle (0.08)
            (uk1) circle (0.08);
    
    \draw[dashed, blue, thick] (u11) .. controls (a) and (b) .. (uextra121) 
    -- (u21) .. controls (c) and (d) .. (uextra2k1) -- (uk1) .. controls  (e) and (f) .. (u11);
    
    \draw[dashed, brown, thick] (u11) .. controls (-2.25, 1.25) and (0.9, 3) .. (u21) .. controls (8.5,4.75) and (7,2) .. (6.5, 1.25) .. controls (6.5, 1) and (5.25, 0.3) .. (uk1) -- (u11);
    
    \draw[dashed, red, thick] (u11) .. controls (-1.5, 1.5) and (-0.5, 3.5) .. (1.5, 4.5) 
    -- (u21) .. controls (6.5, 3.5) and (6, 2) .. (6, 1.75) -- (uk1) .. controls  (3, -1) and (-0.25, -0.75) .. (u11);
    	
    \draw[thick] (Cent) circle (1.5cm);
    
    \draw[thick, dotted] (CentR) circle (0.58cm);
    
    \node at (Space1) {{$\Rcal^1$}};
    \node at (Space2) {{$\Rcal^2$}};
    \node at (Spacem) {{$\Rcal^m$}};

    \node at (Sphere) {\large{$\Scal_\rho$}};
    \node at (Sphereplus) {\tiny{$\Scal^+_{\varrho}$}};
    
    \node[below right] at (Centcent) {$\ombf$};
    
    \node[below left] at (u11) {{$\embf_1$}};
    \node[above] at (u21) {{$\embf_2$}};
    \node[below right] at (uk1) {{$\embf_k$}};
    
     \end{scope}
    \end{tikzpicture}
    \vskip -0.2cm
      \captionof{figure}{$\Rcal^1 \times \dots \times \Rcal^m$ (best seen in colors); $\Rcal^u \, \forall \, u \in [m]$ are convex sets with common vertices $\embf_i \, \forall \, i \in [k]$ and enclose the sphere $\Scal_\rho$.}
        \vskip -0.15cm
      \label{fig:geometry}
    \end{minipage}
  \end{minipage}
  \vskip -0.4cm
\end{figure}

While the problem of selecting an appropriate fairness metric has gained prominence in recent years~\cite{hardt2016equality, menon2018cost,zhang2020joint}, it perhaps best understood as a special case of the task of choosing evaluation metrics in machine learning. For instance, when a cost-sensitive predictive model classifies patients into cancer categories~\cite{yang2014multiclass} even without considering fairness, it is often unclear how the cost-tradeoffs be chosen so that they reflect the expert’s decision-making, i.e., replacing expert intuition by quantifiable metrics. The recently proposed {\rm Metric Elicitation} (ME) framework~\cite{hiranandani2018eliciting,hiranandani2019multiclass} provides a solution. ME is a principled framework for eliciting performance metrics using feedback over classifiers from an end user. The motivation behind ME is that employing the performance metrics which reflect user tradeoffs will enable learning models that best capture user preferences~\cite{hiranandani2018eliciting}. As humans are often inaccurate in providing absolute  preferences~\cite{qian2013active}, Hiranandani et al.~\cite{hiranandani2018eliciting} propose to use pairwise comparison queries, where the user (oracle) is asked to compare two classifiers and provide a relative preference. 
Using such queries, ME aims to recover the oracle's metric. Figure~\ref{fig:ME} (reproduced from~\cite{hiranandani2018eliciting}) illustrates the ME framework.

Existing research suggests a fundamental trade-off between algorithmic fairness and performance~\cite{kamishima2012fairness, zafar2017fairness, corbett2017algorithmic, bechavod2017learning, menon2018cost, zhang2020joint}, where in addition to appropriate metrics, the practitioner or policymaker must choose a trade-off operating point between the competing objectives~\cite{zhang2020joint}. To this end, we extend the ME framework from eliciting multiclass classification metrics~\cite{hiranandani2019multiclass} to the task of eliciting \emph{fair} performance metrics from pairwise preference feedback in the presence of multiple sensitive groups. In particular, we elicit metrics that reflect, jointly, the (i) predictive performance evaluated as a weighting of classifier's overall predictive rates, (ii) fairness violation assessed as the discrepancy in predictive rates among groups, and (iii) a trade-off between the predictive performance and fairness violation. Importantly, the elicited metrics are sufficiently flexible to encapsulate and generalize many existing predictive performance and fairness violation measures.

In eliciting group-fair performance metrics, we tackle three new challenges. First, from preference query perspective, the predictive performance and fairness violations are correlated, thus increasing the complexity of joint elicitation.
Second, we find that in order to measure both positive and negative violations, the fair metrics are necessarily non-linear functions of the predictive rates, thus existing results on linear ME~\cite{hiranandani2019multiclass} cannot be applied directly. Finally, as we show, the number of groups directly impacts query complexity. We overcome these  challenges by proposing a novel query efficient procedure that exploits the geometric properties of the set of rates. 

{\bf Contributions.} We consider metrics for algorithmically group-fair classification and propose a novel approach for eliciting predictive performance, fairness violations, and their trade-off point, from expert pairwise feedback.
Our procedure uses binary-search based subroutines and recovers the metric with linear query complexity. Moreover, the procedure is robust to both  finite sample and oracle feedback noise thus is useful in practice. Lastly, our method can be applied either by querying preferences over classifiers or rates. Such an equivalence is crucial for practical applications~\cite{hiranandani2018eliciting, hiranandani2019multiclass}.

\textbf{Notations.} Matrices and vectors are denoted by bold upper case and bold lower case letters, respectively. We denote the inner product of two vectors by $\inner{\cdot}{\cdot}$ and the Hadamard product by $\odot$. The $\ell_2$-norm is denoted by $\norm{\cdot}_2$. For $k \in \Zmbb_+$, we represent the index set $\{1, 2, \cdots , k\}$ by $[k]$, and the $(k-1)$-dimensional simplex by $\Delta_k$. Given a matrix $\Ambf$, $\offdiag(\Ambf)$ returns a vector of off-diagonal elements of $\Ambf$ in row-major form. The group membership is denoted by superscripts and coordinates of vectors, matrices, and tuples are denoted by subscripts.

%% file: background.tex
\vspace{-0.4cm}
\section{Background}
\label{sec:preliminaries}

The standard multiclass, multigroup classification setting comprises $k$ classes and $m$ groups with $X \in \Xcal$, $G \in [m]$ and $Y \in [k]$ representing the input, group membership, and output random variables, respectively. The groups are assumed to be disjoint and known apriori~\cite{hardt2016equality, kleinberg2017inherent}. We have access to a dataset $\{(\xmbf, g, y)_i\}_{i=1}^n$ of size $n$, generated \emph{iid} from a distribution $ \Pmbb(X, G, Y)$. 

\emph{Group-specific rates:} 
We consider separate (randomized) classifiers $h^g : \Xcal \rightarrow \Delta_k$ for each group $g$, and use 
 $\Hcal^g = \{h^g : \Xcal \rightarrow \Delta_k\}$
 to denote the set of all classifiers for group $g$. 
The group-specific rate matrix $\Rmbf^g(h^g, \Pmbb) \in \Rmbb^{k \times k}$ for a classifier $h^g$ is given by:
\vspace{-0.05cm}
\begin{align}
	R^g_{ij}(h^g, \Pmbb) \coloneqq \Pmbb(h^g = j | Y = i, G= g )  \quad \text{for} \; i, j \in [k].
	\label{eq:components}
\end{align}

Since the diagonal entries of the rate matrix can be written in terms of the off-diagonal entries: \vspace{-0.15cm}
\begin{equation}
    R_{ii}^g(h^g, \Pmbb) = 1 - \sum\nolimits_{j=1,j\neq i}^k R_{ij}^g(h^g, \Pmbb),
    \label{eq:decomp}
\end{equation}
any rate matrix is uniquely represented by its $q \coloneqq (k^2 - k)$ off-diagonal elements  as a vector $\rmbf^g(h^g, \Pmbb) = \offdiag(\Rmbf^g(h^g, \Pmbb))$. So we  will  interchangeably refer to the rate matrix as a \emph{`vector of rates'}. 
The feasible set of rates associated with a group $g$ is denoted by $\Rcal^g = \{\rmbf^g(h^g, \Pmbb) \,:\, h^g \in \Hcal^g \}$.  For clarity, we will suppress the dependence on $\Pmbb$ and $h^g$ if it is clear from the context.

\emph{Overall rates:} We define the overall classifier $h : (\Xcal, [m]) \rightarrow \Delta_k$ by $h(\xmbf, g) \coloneqq h^g(\xmbf)$ and denote its tuple of group-specific rates by:
$$\rmbf^{1:m} \coloneqq  (\rmbf^1, \dots, \rmbf^m) \in \Rcal^1 \times \dots \times \Rcal^m =: \prodRcal.$$ 
This tuple allows us to measure the fairness violation across groups. The fairness violation is believed to be in trade-off with the predictive performance~\cite{kamishima2012fairness, bechavod2017learning, menon2018cost}.  The latter is measured using the overall rate matrix of the classifier $h$:
\vspace{-0.15cm}
\begin{align*}
R_{ij} \coloneqq \Pmbb(h=j|Y=i) 
= \sum\nolimits_{g=1}^m t_{i}^gR_{ij}^g,
    \numberthis \label{eq:overallrate}
\end{align*}
\vskip -0.3cm
where $ t^g_{i} \coloneqq \Pmbb(G=g|Y=i)$ is the prevalence of group $g$ within class $i$. For an overall classifier $h$, the \emph{`vector of rates'} $\rmbf = \offdiag(\Rmbf)$ can be conveniently written in terms of its group-specific tuple of rates as $\rmbf =  \sum_{g=1}^m \bm{\tau}^g \odot \rmbf^g$, where $\bm{\tau}^g \coloneqq \offdiag([\tmbf^g \; \tmbf^g \dots \tmbf^g])$. 

{\em Fairness violation measure:} The (approximate) fairness of a classifier is often determined by the `discrepancy' in rates across different groups e.g. \emph{equalized odds}~\cite{hardt2016equality, barocas2017fairness}. So given two groups $u, v\in[m]$, we define the discrepancy in their rates as:
\vspace{-0.1cm}
\begin{equation}
    \dmbf^{uv} \coloneqq \vert \rmbf^{u} - \rmbf^{v} \vert.
    \label{eq:discrepancy}
\end{equation}
\vskip -0.3cm
Since there are $m$ groups, the number of \emph{discrepancy vectors} are $\tiny{{m\choose 2}}$ . 

\vspace{-0.15cm}
\subsection{Fair Performance Metric}
\label{ssec:metric}
\vskip -0.1cm
We aim to elicit a general class of metrics, which recovers and generalizes existing fairness measures, based on trade-off between predictive performance and fairness violation~\cite{kamishima2012fairness, hardt2016equality, chouldechova2017fair, bechavod2017learning, menon2018cost}.
Let $\phi : [0, 1]^{q}  \rightarrow \Rmbb$ be the cost of overall misclassification (aka.\ predictive performance) and $\varphi : [0, 1]^{m \times q} \rightarrow \Rmbb$ be the fairness violation cost for a classifier $h$ determined by the overall rates $\rmbf(h)$ and group discrepancies $\{\dmbf^{uv}(h)\}_{u,v = 1, v >u}^m$, respectively. 
Without loss of generality (wlog), we assume the metrics $\phi$ and $\varphi$ are costs. 
Moreover, the metrics 
are scale invariant as global scale does not affect the learning problem~\cite{narasimhan2015consistent}; hence let $\phi : [0, 1]^{q}  \rightarrow [0,1]$ and $\varphi : [0, 1]^{m\times q}  \rightarrow [0,1]$.

\bdefinition Fair Performance Metric: Let $\phi$
and $\varphi$ be monotonically increasing linear functions of overall rates and group discrepancies, respectively. The fair metric $\Psi$ is a trade-off between $\phi$ and $\varphi$. In particular, given $\ambf \in \Rmbb^q, \ambf \geq 0$ (misclassification weights), a set of vectors $\Bmbf \coloneqq \{\bmbf^{uv} \in \Rmbb^q, \bmbf^{uv}\geq0\}_{u, v=1, v>u}^m$ (fairness violation weights), and a scalar $\lambda$ (trade-off) with
\vspace{-0.1cm}
\begin{align*}
    \Vert \ambf \Vert_2 = 1, \quad \quad \sum\nolimits_{u,v=1, v>u}^{m} \Vert \bmbf^{uv} \Vert_2 = 1, \quad \quad 0 \leq \lambda \leq 1, \numberthis
    \label{eq:scaleinvariance}
\end{align*}
\vskip -0.1cm
(wlog., due to scale invariance), we define the metric $\Psi$ as:
\vspace{-0.1cm}
\begin{align*}
    \Psi(\tupr \,;\, \ambf, \Bmbf, \lambda) \,\coloneqq\,  \underbrace{(1-\lambda)}_{\text{trade-off}}\underbrace{\inner{\ambf}{\rmbf}}_{\phi(\rmbf)} + \lambda \underbrace{\left(\sum\nolimits_{u,v=1,v>u}^{m} \inner{\mathbbm{\bmbf}^{uv}}{\dmbf^{uv}}\right)}_{\varphi(\tupr)}
    \numberthis \label{eq:linmetric}.
\end{align*}
\vskip -0.1cm
\label{def:linear}
\edefinition

Examples of the misclassification cost $\phi(\rmbf)$ include cost-sensitive linear metrics~\cite{abe2004iterative}. Many existing fairness metrics for two classes and two groups such as \emph{equal opportunity}~\cite{hardt2016equality}, \emph{balance for the negative class}~\cite{kleinberg2017inherent} \emph{error-rate balance} (i.e., $0.5|r_1^1 - r^2_1| + 0.5|r_2^1 - r^2_2|)$~\cite{chouldechova2017fair}, \emph{weighted equalized odds} (i.e., $b_1|r_1^1 - r^2_1| + b_2|r_2^1 - r^2_2|)$~\cite{hardt2016equality, bechavod2017learning}, etc. correspond to fairness violations of the form $\varphi(\tupr)$ considered above. The combination of $\phi(\rmbf)$ and $\varphi(\tupr)$ as defined in $\Psi(\tupr)$ appears regularly in prior work~\cite{kamishima2012fairness, bechavod2017learning, menon2018cost}. Notice that the metric is flexible to allow different fairness violation costs for different pairs of groups thus capable of enabling reverse discrimination~\cite{opotow1996affirmative}. 
Lastly, while the metric is linear with respect to (wrt.) the discrepancies, it is non-linear wrt.\ the group-wise rates. Hence, standard linear ME algorithm~\cite{hiranandani2019multiclass} cannot be trivially applied for eliciting the metric in Definition~\ref{def:linear}.

\subsection{Fair Performance Metric Elicitation; Problem Statement}
\label{ssec:me}

We now state the problem of \emph{Fair Performance Metric Elicitation (FPME)} and define the associated \emph{oracle query}. The broad definitions follow from Hiranandani et al.~\cite{hiranandani2018eliciting, hiranandani2019multiclass}, extended so the rates and the performance metrics correspond to the multiclass multigroup-fair classification setting.

\bdefinition
[Oracle Query] Given two classifiers $h_1,  h_2$ (equivalent to a tuple of rates $\tupr_1, \tupr_2$ respectively), a query to the Oracle (with metric $\Psi$) is represented by:
\vspace{-0.1cm}
\begin{align}
\Gamma(h_1, h_2) = \Omega\left( \tupr_1, \tupr_2 \right) &= \1[\Psi(\tupr_1) > \Psi(\tupr_2)],
\end{align}
\vskip -0.2cm
where $\Gamma: \Hcal \times \Hcal \rightarrow \{0,1\}$ and $\Omega: \prodRcal \times \prodRcal \rightarrow \{0, 1\}$. In words, the query asks whether $h_1$ is preferred to $ h_2$ (equivalent to whether $\tupr_1$ is preferred to $\tupr_2$), as measured by $\Psi$. 
\label{def:query}
\edefinition

In practice, the oracle can be an expert, a group of experts, or an entire user population. The ME framework can be applied by posing classifier comparisons directly to them via interpretable learning techniques~\cite{ribeiro2016should, doshi2017towards} or via A/B testing~\cite{tamburrelli2014towards}. For example, one may perform A/B testing for an internet-based application by deploying two classifiers A and B and use the population's level of engagement to decide the preference between the two classifiers. For other applications, intuitive visualizations of the predictive rates for two different classifiers (see e.g.,  \cite{zhang2020joint,beauxis2014visualization}) can be used to ask preference feedback from a group of domain experts. 

We emphasize that the metric $\Psi$ used by the  oracle is unknown to us and can be accessed only through queries to the oracle. Since the metrics we consider are functions of rates, comparing two classifiers on a metric is equivalent to comparing their corresponding rates. Henceforth, we will denote  any query to the oracle by a pair of rates $(\rmbf_1^{1:m}, \rmbf_2^{1:m})$.
Also, whenever we refer to an oracles's dimension, we are referring to the dimension of its rate arguments. For instance,  we will consider the oracle in Definition~\ref{def:query} to be of dimension $m\times q$. Next, we formally state the FPME problem.

\bdefinition [Fair Performance Metric Elicitation with Pairwise Queries (given $\{(\xmbf,g,y)_i\}_{i=1}^n$)] Suppose that the oracle's (unknown) performance metric is $\Psi$. Using oracle queries of the form $\Omega(\hat \rmbf_1^{1:m}, \hat \rmbf_2^{1:m})$, where $\hat \rmbf_1^{1:m}, \hat \rmbf_2^{1:m}$ are the estimated rates from samples, recover a metric $\hPsi$ such that $\Vert \Psi - \hPsi \Vert < \omega$ under a  suitable norm $\Vert \cdot \Vert$ for sufficiently small error tolerance $\omega > 0$.
\label{def:me}
\edefinition
Similar to the standard metric elicitation problems~\cite{hiranandani2018eliciting, hiranandani2019multiclass}, the performance of FPME is evaluated both by the fidelity of the recovered metric and the query complexity. As done in decision theory literature~ \cite{koyejo2015consistent, hiranandani2018eliciting}, we present our FPME solution by first assuming access to population quantities such as the population rates $\tupr(h, \Pmbb)$, and then discuss how elicitation can be performed from finite samples, e.g., with empirical rates $\hat \rmbf^{1:m}(h, \{(\xmbf,g,y)_i\}_{i=1}^n))$.

\subsection{Linear Performance Metric Elicitation -- Warmup}
\label{ssec:mpme}

We revisit the Linear Performance Metric Elicitation (LPME) procedure from~\cite{hiranandani2019multiclass}, which we will use as as a subroutine to elicit fair performance metrics. The LPME procedure assumes an enclosed sphere $\Scal \subset \Zcal$, where $\Zcal$ is the $q$-dimensional space of classifier statistics that are feasible, i.e., can be achieved by some classifier. It also assumes access to a $q$-dimensional oracle $\Omega'$ whose scale invariant linear metric is of the form $\xi(\zmbf) \coloneqq \inner{\ambf}{\zmbf}$ with $\Vert \ambf \Vert_2=1$, analogous to the misclassification cost in Definition~\ref{def:linear}. Analogously, the oracle queries are of the type $\Omega'( \zmbf_1, \zmbf_2 ) \coloneqq \1[\xi(\zmbf_1) > \xi(\zmbf_2)]$.

When the number of classes $k = 2$, LPME elicits the coefficients $\ambf$ using a simple one-dimensional binary search. When $k > 2$, LPME performs binary search in each coordinate while keeping the others fixed, and performs this in a coordinate-wise fashion until convergence. By restricting this coordinate-wise binary search procedure to posing queries from within a sphere $\Scal$, LPME can be equivalently seen as minimizing a strongly-convex function and shown to converge to a solution $\ambfhat$ close to $\ambf$. Specifically, the algorithm
takes the query space $\Scal\subset\Zcal$, binary-search tolerance $\epsilon$, and the oracle $\Omega'$ as input, and by querying $O(q\log(1/\epsilon))$ queries recovers $\ambfhat$ with $\Vert \ambfhat \Vert_2=1$  such that $\Vert \ambf - \ambfhat \Vert_2 \leq O(\sqrt{q}\epsilon)$ (Theorem~2 in~\cite{hiranandani2019multiclass}). We provide details of the LPME procedure in Algorithm~\ref{alg:slme} (Appendix~\ref{append:sec:slme}) for completeness and summarize the discussion with the following remark. 

\bremark
Given a $q$-dimensional space $\Zcal$ enclosing a sphere $\Scal\subset \Zcal$ and an oracle $\Omega'$ with linear  metric $\xi(\zmbf)\coloneqq\inner{\ambf}{\zmbf}$, the LPME algorithm (Algorithm~\ref{alg:slme}, Appendix~\ref{append:sec:slme}) provides an estimate $\ambfhat$ with $\Vert \ambfhat \Vert_2=1$ such that the estimated slope is close to the true slope, i.e.,  $\sfrac{{a}_i}{{a}_j} \approx \sfrac{\hat a_i}{\hat a_j} \; \forall \; i, j\in [q]$. 
\label{rm:ratio}
\eremark
Note that the algorithm estimates the direction of the coefficient vector, not its magnitude. 

%% file: confusion.tex
\vspace{-0.1cm}
\section{Geometry of the product set $\prodRcal$}
\label{sec:confusion}
\vskip -0.2cm
The LPME  procedure described above works with rate queries of dimension $q$. We would like to use this procedure to elicit the fair metrics in Definition~\ref{def:linear} defined on tuples of dimension $m\times q$. So to make use of LPME, we restrict our queries to a $q$-dimensional sphere $\Scal$ which is common to the feasible rate region $\Rcal^g$ for each group $g$, i.e.\ to a sphere in the intersection $\Rcal^1\cap\ldots\cap\Rcal^m$. We show now that such a sphere does indeed exist under a mild assumption.

\bassumption
For all groups, the conditional-class distributions are not identical, i.e., $\forall\;g\in[m], \forall \;i\neq j, \, \Pmbb(Y=i|X, G=g) \neq \Pmbb(Y=j|X, G=g).$ In other words, there is some non-trivial signal for classification for each group.
\label{as:clsconditional}
\eassumption

Let $\embf_i \in \{0,1\}^q$ be the rate profile for a trivial classifier that predicts class $i$ on all inputs. Note that these trivial classifiers evaluate to the same rates $\embf_i$ irrespective of which group we apply them  to.
\bproposition
[Geometry of $\prodRcal$; Figure~\ref{fig:geometry}] For any group $g\in [m]$, the set of confusion rates $\Rcal^g$ is convex, bounded in $[0, 1]^{q}$, and has vertices $\{\embf_i\}_{i=1}^k$. The intersection of group rate sets $\Rcal^1 \cap \dots \cap \Rcal^m$ is convex and always contains the rate $\ombf = \tfrac{1}{k} \tiny{\sum_{i=1}^k \embf_i}$ in the interior, which is associated with the uniform random classifier that predicts each class with equal probability. 
\label{prop:C}
\eproposition
\vskip -0.1cm
Since $\Rcal^1 \cap \dots \cap \Rcal^m$ is convex and always contains a point $\ombf$ in the interior, we can make the following remark (see Figure~\ref{fig:geometry} for an illustration).

\bremark[Existence of common sphere $\Scal_\rho$] There exists a $q$-dimensional sphere $\Scal_\rho \subset \Rcal^1 \cap \dots \cap \Rcal^m$ of non-zero radius $\rho$ centered at $\ombf$. Thus, any rate $\smbf \in\Scal_\rho$ is feasible for all groups, i.e., $\smbf$ is achievable by some classifier $h^g$ for all groups $g \in [m]$.
\label{remark:sphere}
\eremark

A method to obtain $\Scal_\rho$ with suitable radius $\rho$ from~\cite{hiranandani2019multiclass} is discussed in Appendix~\ref{append:ssec:sphere}. From Remark~\ref{remark:sphere}, we observe that any tuple of group rates $\rmbf^{1:m} = (\smbf^1, \ldots, \smbf^m)$ chosen from $\Scal_\rho \times \ldots \times \Scal_\rho$ is achievable for some choice of group-specific classifiers $h^1, \ldots, h^m$. Moreover, when two groups $u, v$ are assigned the same rate profile $\smbf \in \Scal_\rho$, the fairness discrepancy $\dmbf^{uv} = \bm{0}$. We will exploit these observations in the elicitation strategy we discuss next. 

%% file: elicitation.tex
\section{Metric Elicitation}
\label{sec:me}
\vskip -0.2cm

\begin{figure}[t]
\begin{minipage}[t]{\textwidth}
  \centering \hspace{-0.5em}
   \begin{minipage}[t]{.50\textwidth}
  \vspace{-0.25cm}
  \hspace{-0.5cm}
     \centering
    \includegraphics[scale=0.53]{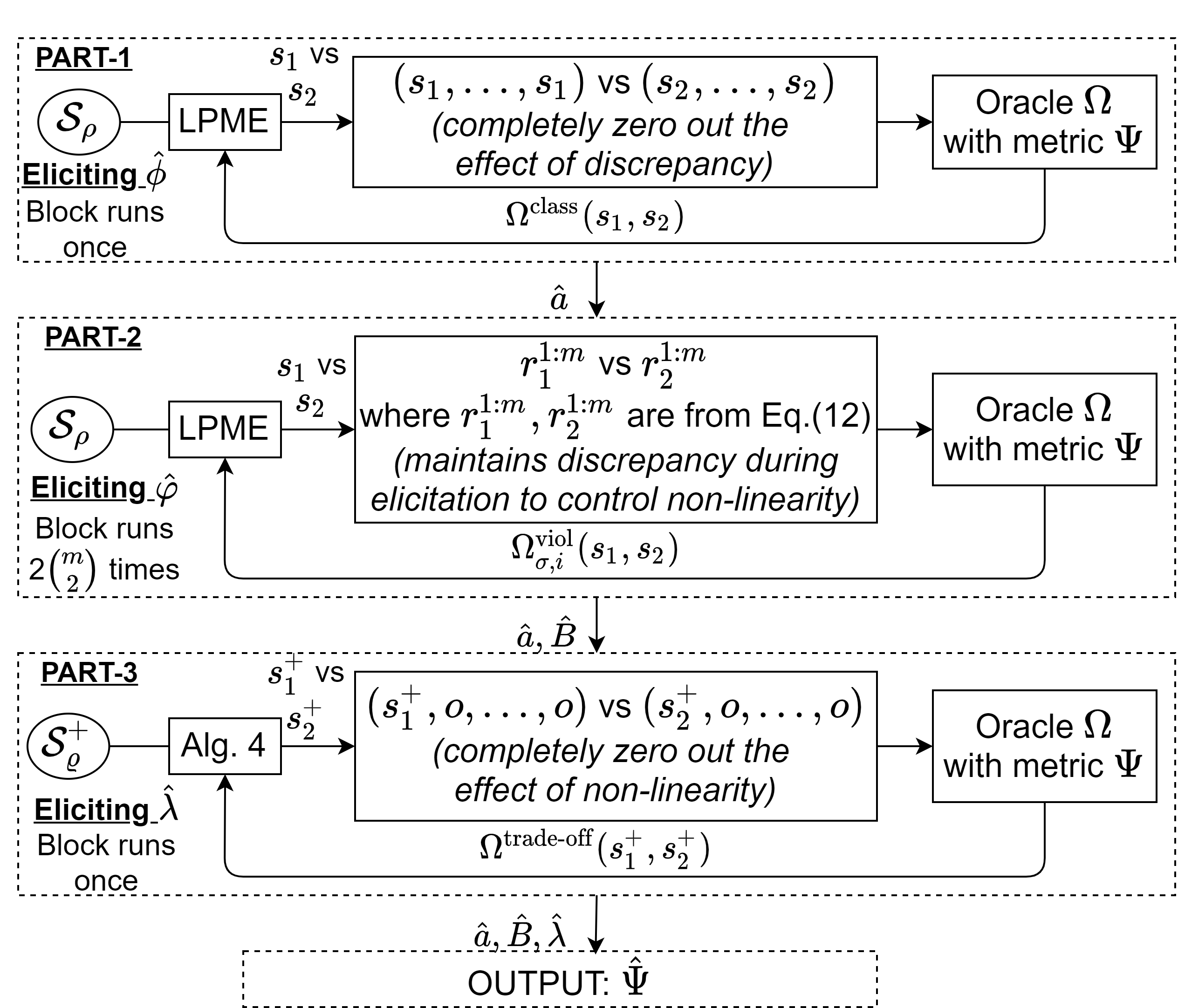}
    \vskip 0.44cm
    \captionof{figure}{Workflow of the FPME procedure.}
    \label{fig:workflow}
  \end{minipage} \hspace{0em}
  \begin{minipage}[t]{.46\textwidth}
  \vspace{0pt}
     \centering
\fbox{\parbox[t]{1.0\textwidth}{\small{\underline{\bf Algorithm \hypertarget{alg:f-me}{\hyperlink{alg:f-me}{1}}: FPM Elicitation}\normalsize}    \\
\small
\textbf{Input:} Query spaces $\Scal_{\rho}$, $\Scal_{\varrho}^+$, 
search tolerance $\epsilon > 0$, and oracle $\Omega$\\
1: \text{ \ }$\ambfhat \leftarrow$ LPME$(\Scal_\rho, \epsilon, \Omega^{\text{class}})$\\
2: \text{ \ }\textbf{If} \, $m==2$\\ 
3: \text{ \ }\text{ \ \ \ }$\breve\fmbf\leftarrow$LPME$(\Scal_\rho, \epsilon, \Omega_1^{\text{viol}})$\\
4: \text{ \ }\text{ \ \ \ }$\tilde\fmbf\leftarrow$LPME$(\Scal_\rho, \epsilon, \Omega_2^{\text{viol}})$\\
5: \text{ \ }\text{ \ \ \ }$\bmbfhat^{12} \leftarrow $ normalized solution from~\eqref{eq:bsolm2}\\
6: \text{ \ }\textbf{Else} Let $\Lcal \leftarrow \varnothing$\\
7: \text{ \ }\text{ \ \ \ }\textbf{For} \, $\sigma \in \Mcal$ \textbf{do}\\
8: \text{ \ \ \ }\text{ \ \ \ } $\breve\fmbf^{\sigma}\leftarrow$LPME$(\Scal_\rho, \epsilon, \Omega_{\sigma, 1}^{\text{viol}})$\\
9: \text{ \ \ \ }\text{ \ \ \ } $\tilde\fmbf^{\sigma}\leftarrow$LPME$(\Scal_\rho, \epsilon, \Omega_{\sigma, k}^{\text{viol}})$\\
10: \text{ \ \ \ }\text{ \ \ } Let $\ell^\sigma$ be Eq.~\eqref{eq:midsolb}, $\Lcal \leftarrow \Lcal \cup \{\ell^\sigma\}$\\
11:  \text{ \ \ \ \ }$\Bmbfhat \leftarrow $ normalized solution~\eqref{eq:bsol} using $\Lcal$\\
12: \text{ \ }$\hat \lambda \leftarrow$ Algorithm~\ref{alg:lambda} $(\Scal_{\varrho}^+, \epsilon, \Omega^{\text{trade-off}})$ \\
\textbf{Output:} $\ambfhat, \Bmbfhat, \hat \lambda$ 
\normalsize 
\vspace{-0.15em}
}}
\label{alg:linear}
  \end{minipage}
  \label{fig:twoalg}
\end{minipage}
\end{figure}

We have access to an oracle whose  (unknown) metric $\bPsi$ given 
in Definition~\ref{def:linear} is parameterized by $(\ambfbar, \Bmbfbar, \lambdabar)$. The proposed FPME framework for eliciting the oracle's metric is presented in Figure~\ref{fig:workflow} and is summarized in Algorithm~\hyperlink{alg:f-me}{1}. The  procedure has three parts executed in sequence: (a) eliciting the misclassification cost $\bphi(\rmbf)$ (i.e., $\ambfbar$), (b) eliciting the fairness violation $\bvarphi(\tupr)$ (i.e., $\Bmbfbar$), and (c) eliciting the trade-off between the misclassification cost and fairness violation (i.e., $\lambdabar$). For simplicity, we will suppress the coefficients $(\ambfbar, \Bmbfbar, \lambdabar)$ from the notation $\Psi$ whenever it is clear from context.

Notice that the metric $\Psi$ is \textit{piece-wise linear} in its coefficients.
So our high level idea  is to restrict the queries we pose to the oracle to lie within regions where the metric $\Psi$ is  linear, so that we can then employ the LPME subroutine to elicit the corresponding linear coefficients. We will show for each of the three components (a)--(c), how we can identify regions in the query space where the metric is linear and apply the LPME procedure (or a variant of it). By restricting the query inputs to those regions, we will essentially be converting the $(m\times q)$-dimensional oracle $\Omega$ in Definition~\ref{def:query} into an equivalent  $q$-dimensional oracle that compares  rates $\smbf_1, \smbf_2$ from the common sphere $ \Scal_\rho \subset \Rcal^1\cap\cdots\cap\Rcal^m$. We first discuss our approach assuming the oracle has no \emph{feedback} noise,  and later in Section~\ref{sec:guarantees} show that our approach is robust to noisy feedback and provide query complexity guarantees.

\vspace{-0.1cm}
\subsection{Eliciting the Misclassification Cost $\bphi(\rmbf)$: {Part 1 in Figure~\ref{fig:workflow} and Line 1 in Algorithm~1}}
\label{ssec:elicitphi}

To elicit the misclassification cost coefficients $\ambfbar$, we will query from a region of the query space where the fairness violation term in the metric is zero. Specifically, we will query group rate profile of the form $\rmbf^{1:m} = (\smbf, \dots, \smbf)$, where $\smbf$ is a $q$-dimensional rate from the common sphere $\Scal_\rho$. For these group rate profiles, the metric $\Psi$ simply evaluates to the linear misclassification term, i.e.:
$$\bPsi(\smbf, \dots, \smbf) =  (1-\lambdabar)\inner{\ambfbar}{\smbf}.$$
So given a pair of group rate profiles $\rmbf_1^{1:m} = (\smbf_1, \dots, \smbf_1)$ and $\rmbf_2^{1:m} = (\smbf_2, \dots, \smbf_2)$, where $\smbf_1, \smbf_2 \in \Scal_\rho$, the oracle's response will essentially compare $\smbf_1$ and $\smbf_2$ on the linear metric $(1-\lambdabar)\inner{\ambfbar}{\smbf}$. Hence, we estimate the coefficients $\ambfbar$ by applying LPME over the $q$-dimensional sphere $\Scal_\rho$ with a modified oracle $\Omega^{\text{class}}$ which takes a pair of rate profiles $\smbf_1$ and $\smbf_2$ from $\Scal_\rho$ as input, and responds with:
\vspace{-0.1cm}
\[
\Omega^{\text{class}}(\smbf_1, \smbf_2) \,=\, \Omega((\smbf_1, \dots, \smbf_1),\, (\smbf_2, \dots, \smbf_2)).
\]
This is decribed in line~1 of Algorithm~\hyperlink{alg:f-me}{1}, which applies the LPME subroutine with query space $\Scal_\rho$, binary search tolerance $\epsilon$, and the oracle $\Omega^{\text{class}}$.  From Remark~\ref{rm:ratio}, this subroutine returns a coefficient vector $\fmbf$ with $\Vert \fmbf \Vert_2=1$ such that:

\begin{equation}
    \frac{(1-\lambdabar)a_i}{(1-\lambdabar)a_j} = \frac{f_i}{f_j} \implies \frac{a_i}{a_j} = \frac{f_i}{f_j}.
\end{equation}

By setting  $\ambfhat = \fmbf$, we recover the classification coefficients independent of the fairness violation coefficients and trade-off parameter. See part 1 in Figure~\ref{fig:workflow} for further illustration. 

\subsection{Eliciting the Fairness Violation $\bvarphi(\tupr)$: Part 2 in Figure~\ref{fig:workflow} and lines 2-11 in Algorithm~1}
\label{ssec:elicitvarphi}

We now discuss eliciting the fairness term $\bvarphi(\tupr)$. We will first discuss the special case of  $m = 2$ groups and later discuss how the proposed procedure can be extended to handle multiple groups. 

\vspace{-0.2cm}
\subsubsection{Special Case of $m=2$: Lines 2-5 in Algorithm~1}
\label{ssec:elicitvarphim2}

Recall from Definition \ref{def:linear} that in the violation term, we measure the group discrepancies using the \textit{absolute} difference between the group rates, i.e.\ $\dmbf^{12} = \vert \rmbf^{1} - \rmbf^{2} \vert$. If we restrict our queries to only those rate profiles $\rmbf^{1:2}$ for which the difference in each coordinate of $\rmbf^{1} - \rmbf^{2}$ is either always positive or always negative, then we can treat the violation term as a linear metric within this region and apply LPME to estimate the associated coefficients. 

To this end, we pose to the oracle queries of the form $\rmbf^{1:2} = (\smbf, \embf_i),$ 
where we assign to group 1 a rate profile $\smbf$ from the common sphere $\Scal_\rho$, and to 
group 2 the  rate profile $\embf_i \in \{0,1\}^q$ for some $i$. Remember that $\embf_i$ is a rate vector associated with a trivial classifier which predicts class $i$ on all inputs, and is therefore a binary vector.
Since we know whether an entry of $\embf_i$ is either a 0 or a 1,  we can decipher the signs of  each entry of the difference vector $\smbf - \embf_i$. 
Hence for group rate profiles of the above form, the metric $\Psi$ can be written as a linear function in $\smbf$:

\begin{align*}
&\bPsi(\smbf, \embf_i) =  
\inner{(1-\lambdabar)\ambfbar \odot (\bm{1}-\bm{\tau}^{2}) + \lambdabar \wmbf_i \odot 
\bmbfbar^{12}}{\smbf} + c_i,
\numberthis \label{eq:metricbrichm2}
\end{align*}
where $\wmbf_i \coloneqq 1 - 2\embf_i$ tells us the sign of each entry of $\smbf - \embf_i$,  $c_i$ is a constant, and we have used the fact that $\taumbf^1 = \bm{1} - \taumbf^2$. Fixing a class $i$, we then apply LPME over the $q$-dimensional sphere $\Scal_\rho$ with a modified oracle $\Omega^{\text{viol}}_i$ which takes a pair of rate profiles $\smbf_1,\smbf_2 \in \Scal_\rho$ as input and responds with:
\begin{equation}
\Omega^{\text{viol}}_i(\smbf_1, \smbf_2) \,=\, \Omega((\smbf_1, \embf_i), (\smbf_2, \embf_i)).
\label{eq:parvarphim2}
\end{equation}
\vskip -0.1cm
One run of LPME with oracle $\Omega^{\text{viol}}_1$ results in $q-1$ independent equations. In order to elicit a $q$-dimensional vector \scl{$\bmbf^{12}$}, we must run LPME again with oracle $\Omega^{\text{viol}}_2$. This is described in lines 3 and 4 of Algorithm~\hyperlink{alg:f-me}{1}. 
The LPME calls provide us with two slopes $\breve \fmbf, \tilde \fmbf$ such that $\Vert \breve \fmbf \Vert_2= \Vert \tilde \fmbf \Vert_2=1$ from which it is easy to obtain the fairness violation weights:
\vspace{-0.1cm}
\begin{align*}
    \bmbfhat^{12} = \frac{\tilde \bmbf^{12}}{\Vert \tilde \bmbf^{12} \Vert_2}, \quad \text{with} \quad
    \tilde \bmbf^{12} = \wmbf_1 \odot \left[ \delta\breve \fmbf - \ambfhat\odot(\bm{1} - \taumbf^{2})  \right], 
    \numberthis \label{eq:bsolm2}
\end{align*}
\vskip -0.1cm
where $\delta$ is a scalar depending on the known entities $\taumbf^{12}, \ambfhat, \breve \fmbf^{12}, \tilde \fmbf^{12}$. The derivation is provided in Appendix~\ref{append:sssec:elicitvarphim2}. Because $\bvarphi$ is scale invariant (see Definition~\ref{def:linear}), the normalized solution \scl{$\bmbfhat^{12}$} is independent of the true trade-off $\lambdabar$ and depends only on the previously elicited vector $\ambfhat$.

\subsubsection{General Case of $m>2$: Lines 6-11 in Algorithm~1}
\label{ssec:elicitvarphim}

We briefly outline the elicitation procedure for $m>2$ groups, with details in Appendix~\ref{append:sssec:elicitvarphi}. Let $\Mcal$ be a set of subsets of the $m$ groups such that each element $\sigma \in \Mcal$ and $[m] \setminus \sigma$  partition the set of $m$ groups. 
We will later discuss how to choose $\Mcal$ for efficient elicitation. Similar to the two-group case, we pose  queries $\rmbf^{1:m}$ where to a subset of groups $\sigma \in \Mcal$, we assign the trivial rate vector $\embf_i$ and to the rest $[m] \setminus \sigma$ groups, we assign a point $\smbf$ from the common sphere $\Scal_\rho$. Observe that within this query region, the metric $\Psi$ is linear in its inputs. So for a fixed partitioning of groups defined by $\sigma$, we apply LPME with a query space $\Scal_\rho$ using the modified $q$-dimensional oracle:
\vspace{-0.1cm}
\begin{equation}
\Omega^{\text{viol}}_{\sigma,i}(\smbf_1, \smbf_2) = \Omega(\rmbf_1^{1:m}, \rmbf_2^{1:m})
~~\text{where}~~ \rmbf_1^g = 
\begin{cases}
    \embf_i & \text{if } g \in \sigma\\
    \smbf_1 & \text{o.w. }
\end{cases}
~~\text{and}~~ \rmbf_2^g = 
\begin{cases}
    \embf_i & \text{if } g \in \sigma\\
    \smbf_2 & \text{o.w. }
\end{cases}.
\label{eq:parvarphi}
\end{equation}
\vskip -0.1cm
As described in lines 8 and 9 of the algorithm, we repeat this twice fixing class $i$ to 1 and $k$. The guarantees for LPME then give us the following relationship between coefficients $\bmbfbar^{uv}$ we wish to elicit and the already elicited coefficient $\hat{\ambf}$:
\begin{align*}
    \sum\nolimits_{u, v} \1\left[|\{u,v\}\cap \sigma|=1\right] \tilde \bmbf^{uv} = \wmbf_1 \odot \left[ \delta^{\sigma} \breve \fmbf^{\sigma} - \ambfhat\odot (\bm{1} - \taumbf^{\sigma}) \right],  \numberthis 
    \label{eq:midsolb}
\end{align*}
where \scl{$\taumbf^{\sigma} = \sum_{g\in \sigma}\bm{\tau}^{g}$} and $\tilde \bmbf^{uv} \coloneqq \lambdabar\bmbfbar^{uv}/(1 - \lambdabar)$ is a scaled version of the true (unknown) $\bmbfbar^{uv}$.
Since we need to estimate $\tiny{{m\choose 2}}$ coefficients, we repeat the above procedure for $\tiny{{m\choose 2}}$ partitions of the groups defined by $\sigma$  and get a system of $\tiny{{m\choose 2}}$ linear equations. We may choose any $\Mcal$ of size $\tiny{{m\choose 2}}$ so that the equations are independent. From the solution to these equations, we recover $\tilde \bmbf^{uv}$'s, which we normalize to get estimates of the final fairness violation weights:
\vspace{-0.1cm}
\begin{equation}
\bmbfhat^{uv} = \frac{\tilde \bmbf^{uv}}{\sum_{u,v=1, v > u}^m \Vert \tilde \bmbf^{uv} \Vert_2} \quad \text{for} \quad u,v \in [m], v>u.
 \label{eq:bsol}
\end{equation}
Thanks to the normalization, the elicited fairness violation weights are independent of the trade-off $\lambdabar$.

\subsection{Eliciting Trade-off $\lambdabar$: Part 3 in Figure~\ref{fig:workflow} and Line 12 in Algorithm~1}
\label{ssec:elicitlambda}
\vskip -0.2cm
Equipped with estimates of the misclassification and fairness violation coefficients $(\hat{\ambfbar}, \hat{\Bmbfbar})$, the 
 final step is to elicit the trade-off $\lambdabar$ between them.  We now show how this can be posed as one-dimensional binary search problem. Suppose we restrict our queries to be of the form $\rmbf^{1:m} = (\smbf^+, \ombf, \ldots, \ombf),$ where for all but the first group, we assign the rate $\ombf$ associated with a uniform random classifier, and for the first group, we assign some rate $\smbf^+$ such that $\smbf^+ \geq \ombf$. For these rate profiles, the group rate difference terms $\rmbf^1 - \rmbf^v = \smbf^+ -\ombf \geq \mathbf{0}$ for all $v \in \{2, \ldots, m\}$, and all the other difference terms are $\mathbf{0}$. As a result, the metric $\Psi$ is linear in the input rate profiles:
\vspace{-0.1cm}
\begin{align*}
        \bPsi(\smbf^+, \ombf, \ldots, \ombf) = \inner{(1-\lambdabar)\taumbf^1\odot\ambfbar + \lambdabar \sum\nolimits_{v=2}^m \bmbfbar^{1v}}{\smbf^+} + c,
         \numberthis \label{eq:metriclambda}
\end{align*}
\vskip -0.2cm
where $c$ is a constant. 
Despite the metric being linear in the identified input region, we cannot directly apply the LPME procedure described in Section \ref{ssec:mpme} to elicit $\lambda$, because we have one parameter to elicit but the input to the metric is $q$-dimensional. Here we propose a slight variant of LPME.

Similar to the original procedure \cite{hiranandani2018eliciting}, we first construct a one-dimensional function $\vartheta$, which takes a guess of the trade-off parameter as input, and outputs the quality of the guess. We show that this function is unimodal and its mode coincides with the oracle's true trade-off parameter $\lambda$. 

\blemma
Let $\Scal_\varrho^+ \subset \Scal_\rho$ be a $q$-dimensional sphere with radius $\varrho < \rho$ such that $\smbf^+ \geq \ombf, \, \forall\, \smbf^+ \in \Scal^+_\varrho$ (see Figure~\ref{fig:geometry}). Assume the estimates $\hat{\ambf}$ and $\hat{\bmbf}^{uv}$'s satisfy a mild regularity condition $\inner{\hat{\ambf}}{\sum_{v=2}^m \hat{\bmbf}^{1v}}\neq 1$. Define a one-dimensional function $\vartheta$ as:
\begin{equation}
\vartheta(\bar{\lambda}) \coloneqq \Psi(\smbf_{\bar{\lambda}}^*, \ombf, \ldots, \ombf),
\label{eq:vartheta}
\end{equation}
where
~\\[-15pt]
\begin{equation}
\smbf^*_{\bar{\lambda}} \,=\, \argmax_{s^+ \in \Scal_\varrho^+}\,\inner{(1-\bar{\lambda})\taumbf^1\odot\hat{\ambfbar} + \bar{\lambda} \sum\nolimits_{v=2}^m \hat{\bmbfbar}^{1v}}{\smbf^+}.
\label{eq:vartheta-max}
\end{equation}
\vskip -0.3cm
Then the function 
$\vartheta$ is strictly quasiconcave (and therefore unimodal) in $\bar{\lambda}$. Moreover, the mode of this function is achieved at the oracle's true trade-off parameter ${\lambda}$.
\label{lm:lambda}
\elemma

For a candidate trade-off $\bar{\lambda}$, 
the function $\vartheta$ first constructs a candidate linear metric based on \eqref{eq:metriclambda}, maximizes this candidate metric over inputs $\smbf^+$, and evaluates the oracle's true metric $\Psi$ at the maximizing rate profile. Note that we cannot directly compute the function $\vartheta$ as it needs the oracle's metric $\Psi$. However, given two candidates for the trade-off parameter $\bar{\lambda}_1$ and $\bar{\lambda}_2$, one can compare the values of $\vartheta(\bar{\lambda}_1)$ and $\vartheta(\bar{\lambda}_2)$ by finding the corresponding maximizers over $\smbf^+$ and querying the oracle to compare them. Because  $\vartheta$ is unimodal, one can use a simple binary search using such pairwise comparisons to find the mode of the function, which we know coincides with the true $\lambda$. We provide an outline of this procedure in Algorithm~\ref{alg:lambda} in Appendix \ref{append:ssec:lambda}, which uses the modified oracle $$\Omega^{\text{trade-off}}(\smbf_1^+, \smbf_2^+) = \Omega((\smbf^+_1, \ombf, \ldots, \ombf),\, (\smbf^+_2, \ombf, \ldots, \ombf))$$ to compare the maximizers in \eqref{eq:vartheta-max}. We also discuss in Appendix \ref{append:ssec:lambda}  how the maximizer in \eqref{eq:vartheta-max} can be computed efficiently. Combining parts 1, 2 and 3 in Figure~\ref{fig:workflow} completes the FPME procedure.

%% file: guarantees.tex
\section{Guarantees}
\label{sec:guarantees}
\vskip -0.4cm
We discuss elicitation guarantees under the following feedback model. 

\bdefinition[Oracle Feedback Noise: $\epsilon_\Omega \geq 0$] For two rates $\tupr_1, \tupr_2 \in \prodRcal$, the oracle responds correctly as long as $|\bPsi(\tupr_1) - \bPsi(\tupr_2)| > \epsilon_\Omega$. Otherwise, it may be incorrect.
\label{def:noise}
\edefinition

In words, the oracle may respond incorrectly if the rates are very close as measured by the metric $\bPsi$. Since deriving the final metric involves offline computations including certain ratios, we discuss guarantees under a regularity assumption that ensures all components are well defined. 

\bassumption
We assume that $1 > c_1 > \lambdabar > c_2 > 0$, $\min_{i}\vert a_i\vert > c_3$, $\min_{i}\vert (1-\lambdabar)a_i {\tau}^{\sigma}_i - \lambdabar w_{ji}
b^{\sigma}_i \vert > c_4 \, \forall\, j\in [q], \sigma \in \Mcal$,  for some $c_1, c_2, c_3, c_4 > 0$, $\rho > \varrho \gg \epsilon_\Omega$, and $\inner{\ambfbar}{\sum_{v=2}^m \bmbfbar^{1v}}\neq 1$. 
\label{as:regularity}
\eassumption

\btheorem 
Given $\epsilon,\epsilon_\Omega\geq 0$, and a 1-Lipschitz fair performance metric $\;\bPsi$ parametrized by $\ambfbar, \Bmbfbar, \lambdabar$, under Assumptions~\ref{as:clsconditional} and~\ref{as:regularity}, Algorithm~1 returns a metric $\hPsi$ with parameters:\vspace{-0.25cm}
\begin{itemize}[itemsep=0pt, leftmargin=1em]
    \item $\ambfhat:$ after $O\left(q\log \tfrac 1 {\epsilon}\right)$ queries such that $\Vert \ambfbar-\ambfhat \Vert_{2}\leq O\left(\sqrt{q}(\epsilon+\sqrt{\epsilon_\Omega/\rho})\right)$.
    \item $\Bmbfhat:$ after $O\left({m\choose 2} q\log \tfrac 1 {\epsilon}\right)$ queries such that $\Vert \text{vec}(\Bmbfbar)-\text{vec}(\Bmbfhat) \Vert_{2}\leq O\left(mq(\epsilon+\sqrt{\epsilon_\Omega/\rho})\right)$, where $\text{vec}(\cdot)$ vectorizes the matrix.
    \item $\lambdahat:$ after $O(\log(\tfrac{1}{\epsilon}))$ queries, with error $\vert \lambdabar - \lambdahat \vert\leq O\left(\epsilon  + \sqrt{\epsilon_\Omega/\varrho} +  \sqrt{m q (\epsilon + \sqrt{\epsilon_\Omega/\rho})/\varrho}\right)$.\vspace{-0.1cm}
\end{itemize}
 \label{thm:error}
\etheorem 
\vskip -0.5cm
We see that the proposed FPME procedure is robust to noise, and its query complexity depends linearly in the number of unknown entities. For instance, line 1 in Algorithm~1 elicits $\ambfhat \in \Rmbf^q$ by posing $\tilde O(q)$ queries, the `for' loop in line~7 of Algorithm~1 runs for $\tiny{m\choose 2}$ iterations, where each iteration
requires $\tilde O(2q)$ queries, and finally line~12 in Algorithm~1 is a simple binary search requiring $\tilde O(1)$ queries. Previous work suggests that linear multiclass elicitation (LPME) elicits misclassification costs ($\phi$) with linear query complexity \cite{hiranandani2019multiclass}. Surprisingly, our proposed FPME procedure elicits a more complex (nonlinear) metric without increasing the query complexity order. Furthermore, since sample estimates of rates are consistent estimators, and the metrics discussed are $1$-Lipschitz wrt.\ rates, with high probability, we gather correct oracle feedback from querying with finite sample estimates $\Omega(\hat{\rmbf}^{1:m}_1, \hat{\rmbf}^{1:m}_2)$ instead of querying with population statistics $\Omega({\rmbf}^{1:m}_1, {\rmbf}^{1:m}_2)$, as long as we have sufficient samples. Apart from this, Algorithm~1 is agnostic to finite sample errors as long as the sphere $\Scal_\rho$ is contained within the feasible region $\Rcal^1 \cap \dots \cap \Rcal^m$. 

%% file: experiments.tex
\vspace{-0.5cm}
\section{Experiments}
\label{sec:extensions}
We first empirically validate the FPME procedure and recovery guarantees in  Section~\ref{ssec:recovery} and then highlight the utility of FPME in evaluating real-world classifiers in Section~\ref{ssec:ranking} .  

\begin{figure*}[t]
	\centering 
	\subfigure{
		{\includegraphics[width=5cm]{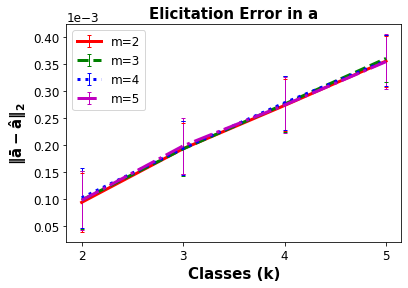}}
		\label{fig:rec_a}
	}\quad
	\subfigure{
		{\includegraphics[width=5cm]{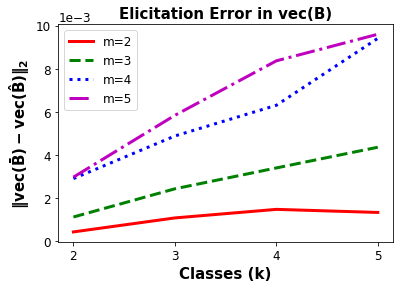}}
		\label{fig:rec_B}
	}\quad
	\subfigure{
		{\includegraphics[width=5cm]{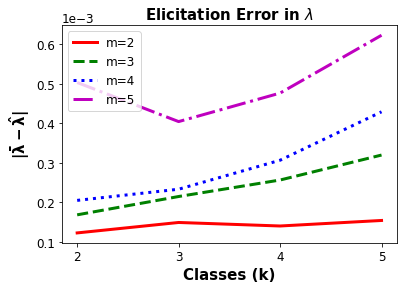}}
		\label{fig:rec_l}
	}
	\vskip -0.4cm
	\caption{Elicitation error in recovering the oracle's metric.}
	\label{fig:recovery}
\end{figure*}

\subsection{Recovering the Oracle's Fair Performance Metric}
\label{ssec:recovery}

Recall that there exists a sphere $\Scal_\rho \subset \Rcal^1 \cap \dots \cap \Rcal^m$ as long as there is a non-trivial classification signal within each group (Assumption~\ref{remark:sphere}). Thus for experiments, we assume access to a feasible sphere $\Scal_\rho$ with $\rho = 0.2$.  We randomly generate 100 oracle metrics each for $k, m \in \{2,3,4,5\}$ parametrized by $\{\ambfbar, \Bmbfbar, \lambdabar\}$. This specifies the query outputs by the oracle for each metric in Algorithm~\hyperlink{alg:f-me}{1}. We then use Algorithm~\hyperlink{alg:f-me}{1} with tolerance $\epsilon = 10^{-3}$ to elicit corresponding metrics parametrized by $\{\ambfhat, \Bmbfhat, \lambdahat\}$. Algorithm~1 makes $1 + 2M$ subroutine calls to LPME procedure and $1$ call to Algorithm~\ref{alg:lambda}. LPME subroutine requires exactly $16(q-1)\log(\pi/2\epsilon)$ queries, where we use 4 queries to shrink the interval in the binary search loop and fix 4 cycles for the coordinate-wise search. Also, Algorithm~\ref{alg:lambda} requires $4\log(1/\epsilon)$ queries.

In Figure~\ref{fig:recovery}, we report the mean of the $\ell_2$-norm between the oracle's metric and the elicited metric. 
Clearly, we elicit metrics that are close to the true metrics. Moreover, this holds true across a range of $m$ and $k$ values demonstrating the robustness of the proposed approach. Figure~\ref{fig:rec_a} shows that the error $\Vert \ambfbar - \ambfhat\Vert_2$ increases only with the number of classes $k$ and not groups $m$. This is expected since $\ambfhat$ is elicited by querying rates that zero out the fairness violation (Section~\ref{ssec:elicitphi}). Figure~\ref{fig:rec_B} verifies Theorem~\ref{thm:error} by showing that $\Vert \text{vec}(\Bmbfbar) - \text{vec}(\Bmbfhat) \Vert_2$ increases with both number of  classes $k$ and groups $m$. In accord with Theorem~\ref{thm:error}, Figure~\ref{fig:rec_l} shows that the elicited trade-off $\hat \lambda$ is also close to the true $\lambdabar$. However, the elicitation error increases consistently with groups $m$ but not with classes $k$. A possible reason may be the cancellation of errors from eliciting $\ambfhat$ and $\Bmbfhat$ separately.

\subsection{Ranking of Classifiers for Real-world Datasets}
\label{ssec:ranking}

One of the most important applications of performance metrics is evaluating classifiers, i.e., providing a quantitative score for their quality which then allows us to choose the best (or best set of) classifier(s). In this section, we discuss how the ranking of plausible classifiers is affected when a practitioner employs default metrics to rank (fair) classifiers instead of the oracle's metric or our elicited approximation. 

We take four real-world classification datasets with $k, m \in \{2,3\}$ (see Table~\ref{append:tab:stats}). 60\% of each dataset is used for training and the rest for testing. We create a pool of 100 classifiers for each dataset by tweaking hyperparameters under logistic regression models~\cite{kleinbaum2002logistic}, multi-layer perceptron models~\cite{pal1992multilayer}, support vector machines~\cite{joachims1999svmlight}, LightGBM models~\cite{ke2017lightgbm}, and fairness constrained optimization based models~\cite{narasimhan2019optimizing}. We compute the group wise confusion rates on the test data for each model for each dataset. We will compare the ranking of these classifiers achieved by competing baseline metrics with respect to the ground truth ranking. 

\begin{table}[t]
\centering
\caption{\small{Dataset statistics; the real-valued regressor in \emph{wine} and \emph{crime} is binned to three classes.}}
\begin{tabular}{|c|ccccc|}
\hline
\textbf{Dataset} & $k$ & $m$ & \textbf{\#samples} & \textbf{\#features} & \textbf{group.feat} \\ 
\hline
default        & 2  & 2  & 30000           &    33    &     gender       \\
adult        &  2 &  3 &    43156       &    74    &    race        \\
wine        &  3 &  2 & 6497          &     13   &  color          \\
crime        &  3 & 3  &    1907       &     99   &      race   \\ 
\hline
\end{tabular}
\vskip -0.2cm
\label{append:tab:stats}
\end{table}

\begin{table}[t]
    \centering
    \caption{\small{Common (baseline) metrics usually deployed to rank classifiers.}}
    \begin{tabular}{|c|cccccccc|}
    \hline 
    \textbf{Name} $\rightarrow$  & {${\hphi\hvarphi\lambdahat}$\_a} & {$\hphi\hvarphi\lambdahat$\_w} & {$\hphi\hvarphi$\_a} & {$\hphi\hvarphi$\_w} & {$\hphi$\_a} & {$\hphi$\_w} & 
    o\_p & 
    o\_f \\ \hline
    $\ambfhat$  & acc. & w-acc. & acc. & w-acc.  & acc. & w-acc. & $\ambfbar$ & - \\ 
    $\Bmbfhat$  & acc. & w-acc. & acc. & w-acc.  & elicit & elicit & - & $\Bmbfbar$ \\ 
    $\lambdahat$ & $0.5$  & w-acc. & elicit & elicit & elicit  & elicit & 0 & 1 \\ 
    \hline
\end{tabular}
    \label{append:tab:baselines}
    \vskip -0.4cm
\end{table}

We generate 100 random oracle metrics $\bPsi$. $\bPsi$'s gives us the ground truth ranking of the above classifiers. We then use our proposed procedure FPME (Algorithm~\hyperlink{alg:f-me}{1}) to recover the oracle's metric. For comparison in ranking of real-world classifiers, we choose a few metrics that are routinely employed by practitioners as baselines (see Table~\ref{append:tab:baselines}). 
The prefixes (i.e. $\hphi, \hvarphi$, or $\lambdahat$) in name of the baseline metrics denote the components that are set to default metrics, and the suffixes (i.e. `a' or `wa') denote whether the assignment is done with \emph{accuracy} (i.e. equal weights) or with \emph{weighted accuracy} (weights are assigned randomly however maintaining the true order of weights as in $\bPsi$). 
For example, $\hphi\hvarphi\lambdahat$\_a  corresponds to the metric where {$\hphi, \hvarphi, \lambdahat$} are set to standard classification accuracy. Similarly, {$\hphi$\_w} denote a metric where the misclassification cost {$\hphi$} is set to weighted accuracy but both $\hvarphi$ and $\lambdahat$ are elicited using Part 2 and Part 3 of the FPME procedure (Algorithm~1), respectively. Assigning weighted accuracy versions is a commonplace since sometimes the order of the costs associated with the types of mistakes in misclassification cost $\bphi$ or fairness violation $\bvarphi$ or preference for fairness violation over misclassification $\lambdabar$ is known but not the actual cost. 
Another example is {$\hphi\hvarphi$\_a} which corresponds to the metric where {$\hphi, \hvarphi$} are set to accuracy and only the trade-off {$\lambdahat$} is elicited using Part 3 of the FPME procedure (Algorithm~1). 
This is similar to prior work by Zhang et al.~\cite{zhang2020joint} who assumed the classification error and fairness violation known, so only the trade-off has to be elicited -- however they also assume direct ratio queries, which can be challenging in practice. Our approach applies much simnpler pairwise preference queries. 
Lastly, o\_p and o\_f represent \emph{only predictive performance} with $\lambda=0$ and \emph{only fairness} with $\lambda=1$, respectively. 

\begin{figure*}[t]
	\centering 
	\subfigure{
		{\includegraphics[width=6cm]{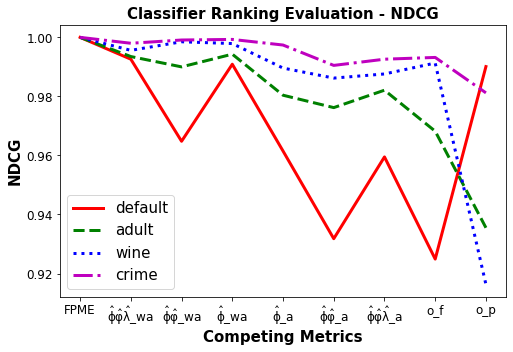}}
		\label{fig:rec_B}
	}\quad\quad
	\subfigure{
		{\includegraphics[width=6cm]{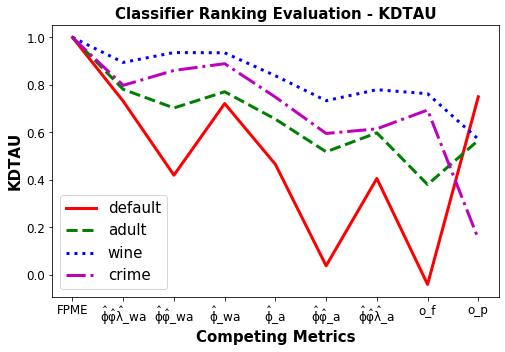}}
		\label{fig:rec_l}
	}
	\vskip -0.4cm
	\caption{Ranking Performance of competing metrics while ranking real-world classifiers.}
	\label{append:fig:ranking}
	\vskip -0.45cm
\end{figure*}

Figure~\ref{append:fig:ranking} shows average NDCG (with exponential gain)~\cite{valizadegan2009learning} and Kendall-tau coefficient~\cite{shieh1998weighted} over 100 metrics $\bPsi$ and their respective estimates by the competing baseline metrics. We see that FPME, wherein we elicit $\hphi, \hvarphi$, and $\lambdahat$ in sequence, achieves the highest possible NDCG and Kendall-tau coefficient. Even though we make some elicitation error in recovery (Section~\ref{sec:extensions}), we achieve almost perfect results while ranking the classifiers. To connect to practice, this implies that when given a set of classifiers, ranking based on elicited metrics will align most closely to ranking based on the true metric, as compared to ranking classifiers based on default metrics. This is a crucial advantage of metric elicitation for practical purposes. In this experiment, baseline metrics achieve inferior ranking of classifiers in comparison to the rankings achieved by metrics that are elicited using the proposed FPME procedure. Figure~\ref{append:fig:ranking} also suggests that it is beneficial to elicit all three components $(\ambfbar, \Bmbfbar, \lambdabar)$ of the metric in Definition~\ref{def:linear}, rather than pre-define a component and elicit the rest. For the \emph{crime} dataset, some methods also achieve high NDCG values, so ranking at the top is good; however Kendall-tau coefficient is weak which suggests that overall ranking is poor. With the exception  of the \emph{default} dataset, the weighted versions are better than equally weighted versions in ranking. This is expected because in weighted versions, at least order of the preference for the type of costs matches with the oracle's  preferences.

%% file: relatedwork.tex
\vspace{-0.6cm}
\section{Related Work}
\label{sec:relatedwork}
\vskip -0.2cm
Some early attempts to eliciting 
individual fairness metrics~\cite{ilvento2019metric, mukherjee2020two} are distinct from ours – as we are focused on the more prevalent setting of group fairness, yet for which there
are no existing approaches to our knowledge. 
Zhang et al.~\cite{zhang2020joint} propose an approach that elicits only the trade-off between accuracy and fairness using complicated ratio queries. We, on the other hand, elicit classification cost, fairness violation, and the trade-off together as a non-linear function, all using much simpler pairwise comparison queries. Prior work for constrained classification focus on learning classifiers under constraints for fairness~\cite{goh2016satisfying, hardt2016equality, zafar2017constraints,narasimhan2018learning}. We take the regularization view of algorithmic fairness, where a fairness violation is embedded in the metric definition instead of as constraints~\cite{kamishima2012fairness, bechavod2017learning, corbett2017algorithmic, agarwal2018reductions, menon2018cost}. 
From the elicitation  perspective, the closest line of work to ours is Hiranandani et al.~\cite{hiranandani2018eliciting, hiranandani2019multiclass}, who propose the problem of ME but solve it only for a simpler setting of classification without fairness. As we move to multiclass, multigroup fair performance ME, we find that the complexity of both the form of the metrics and the query space increases. This results in starkly different elicitation strategy with novel methods required to provide query complexity guarantees. Learning (linear) functions passively using pairwise comparisons is a mature field~\cite{joachims2002optimizing, herbrich2000large, peyrard2017learning}, but these approaches fail to control sample (i.e.\ query) complexity. 
Active learning in fairness~\cite{noriega2019active} is a related direction; however the aim there is to learn a fair classifier based on fixed metric instead of eliciting the metric itself.  

%% file: discussion.tex
\vspace{-0.3cm}
\section{Discussion Points and Future Work}
\label{sec:extensions}
\vspace{-0.1cm}
\bitemize[itemsep=0pt, leftmargin=*]
\item \textbf{Transportability:} Our elicitation procedure is independent of the population $\Pmbb$ as long as there exists a sphere of rates which is feasible for all groups. Thus, any metric that is learned using one dataset or model class (i.e. by estimated $\hat \Pmbb$) can be applied to other applications and datasets, 
as long as the expert believes the context and tradeoffs are the same.
\item \textbf{Extensions.} Our propsal can be modified to leverage the structure in the metric or the groups to further reduce the query complexity. For example, when the fairness violation weights are the same for all pairs of groups,  the procedure in Section~\ref{ssec:elicitvarphim} requires only one partitioning of groups to elicit the metric $\hvarphi$. Such modifications are easy to incorporate. In the future, we  plan to extend our approach to more complex metrics such as linear-fractional functions of rates and discrepancies.
\item \textbf{Limitations of group-fair metrics.} Since the metrics we consider depend on a classifier only through its rates, comparing two classifiers on these metrics is equivalent to comparing
their rates. Unfortunately, with this setup, all the limitations associated with group-fairness definition of metrics apply to our setup as well. For example, we may discard notions of \emph{individual fairness} when only group-rates are considered for comparing classifiers~\cite{binns2020apparent}. Similarly, issues associated with \emph{overlapping groups}~\cite{kearns2018preventing}, \emph{detailed group specification}~\cite{kearns2018preventing}, \emph{unknown or changing groups}~\cite{hashimoto2018fairness, gillen2018online}, \emph{noisy or biased} group information~\cite{wang2020robust}, among others, pose limitations to our proposed setup. We hope that as the first work on the topic, our work will inspire the research community to address many of these open problems for the task of metric elicitation. 
\item \textbf{Optimal bounds.} We conjecture that our query complexity bounds are tight; however, we leave this detail for the future. 
In conclusion, we elicit a more complex (non-linear) group fair-metric with the same query complexity order as standard classification linear elicitation procedures~\cite{hiranandani2019multiclass}.
\eitemize

%% file: conclusion.tex
\vspace{-0.4cm}
\section{Conclusion}
\label{sec:discussion}
\vskip -0.2cm
We study the space of multiclass, multigroup predicitve rates and propose a novel, provably query efficient strategy to elicit group-fair performance metrics. The proposed procedure only requires pairwise preference feedback over classifiers and and is robust to finite sample and feedback noise.

%% file: supplement.tex
\renewcommand{\thesection}{\Alph{section}}
\setcounter{section}{0}
\newcommand{\pb}{\vspace*{-\parskip}\noindent\rule[0.5ex]{\linewidth}{1pt}}

\begin{appendices}

\section{Linear Performance Metric Elicitation}
\label{append:sec:slme}

As explained in Section~\ref{ssec:mpme}, we use the linear metric elicitation procedure~\cite{hiranandani2019multiclass} as a subroutine in order to elicit a more complicated metric as defined in Definition~\ref{def:linear}. For completeness, we provide the details here.

The linear metric elicitation procedure proposed in~\cite{hiranandani2019multiclass} assumes an enclosed sphere $\Scal \subset \Zcal$, where $\Zcal$ is the $q$-dimensional space of classifier statistics that are feasible, i.e., can be achieved by some classifier. Let the the radius of the sphere $\Scal$ be $\rho$. We extend the linear metric elicitation procedure (Algorithm 2 in~\cite{hiranandani2019multiclass}) to elicit any linear metric (without the monotonicity condition) defined over the space $\Zcal$. This is because in Section~\ref{ssec:elicitvarphi}, we require to  elicit slopes that are not necessarily for monotonic metrics (e.g., see Equation~\eqref{eq:metricbrichm2}). Let the oracle's scale invariant metric be $\xi(\zmbf) \coloneqq \inner{\ambf}{\zmbf}$, such that $\Vert \ambf \Vert_2=1$. Analogously, the oracle queries are $\Omega'( \zmbf_1, \zmbf_2) \coloneqq \1[\xi(\zmbf_1) > \xi(\zmbf_2)]$. We start by outlining a trivial Lemma from~\cite{hiranandani2019multiclass}.

\blemma~\cite{hiranandani2019multiclass}
Let $\xi$ be a linear metric parametrized by $\ambf$ such that $\Vert \ambf \Vert_2=1$, then the unique optimal classifier statistic $\zmbfbar$ over the sphere $\Scal$ is a point on the boundary of $\Scal$ given by $\zmbfbar = \rho \ambf +\ombf$, where $\ombf$  is the center of the sphere $\Scal$. 
\label{lem:spherebayes}
\elemma

Given a linear performance metric, Lemma~\ref{lem:spherebayes} provides a unique point in the query space which lies on the boundary of the sphere $\partial\Scal$. Moreover, the converse also holds true that given a point on the boundary of the sphere $\partial\Scal$, one may recover the linear metric for which the given point is optimal. Thus, in order to elicit a linear metric, Hiranandani et al.~\cite{hiranandani2019multiclass} essentially search for the optimal statistic (over the surface of the sphere) using pairwise queries to the oracle which in turn reveals the true metric. The algorithm is summarized in Algorithm~\ref{alg:slme}. The algorithm also uses the following standard paramterization for the surface of the sphere $\partial\Scal$. 

\textbf{Parameterizing the boundary of the enclosed sphere $\partial \Scal$.} 
Let $\thetambf$ be a ($q-1$)-dimensional vector of angles, where all the angles except the primary angle are in $[0, \pi]$, and the primary angle is in $[0, 2\pi]$. A linear performance metric with $\Vert \ambf \Vert_2=1$ is constructed by setting $a_i = \Pi_{j=1}^{i-1} \sin\theta_j \cos{\theta_i}$ for $i \in [q-1]$ and $a_q = \Pi_{j=1}^{q-1} \sin\theta_j$. By using Lemma~\ref{lem:spherebayes}, the metric's optimal classifier statistic over the sphere $\Scal$ is easy to compute. Thus, varying $\thetambf$ in this procedure, parametrizes the surface of the sphere $\partial\Scal$. We denote this parametrization by $\mu(\thetambf)$, where $\mu: [0, \pi]^{q-2} \times [0, 2\pi] \to \partial \Scal$.

\begin{figure}[t]
\begin{minipage}[h]{\textwidth}
  \centering \hspace{-0.5em}
  \begin{minipage}[h]{.55\textwidth}
     \centering
\fbox{\parbox[t]{1.0\textwidth}{\vspace{0.0cm}\small{\underline{\bf Subroutine \emph{ShrinkInterval}}\normalsize}    \\
\small
\textbf{Input:} Oracle responses for $\Omega'(\zmbfbar^{(c)}, \zmbfbar^{(a)})$,\\
$\Omega'(\zmbfbar^{(d)}, \zmbfbar^{(c)}),$ $\Omega'(\zmbfbar^{(e)}, \zmbfbar^{(d)}), \Omega'(\zmbfbar^{(b)},  \zmbfbar^{(e)})$\\
\textbf{If} \, ($\zmbfbar^{(a)} \succ \zmbfbar^{(c)}$) Set $\theta_j^{(b)} = \theta_j^{(d)}$.\\
\textbf{elseif} \, ($\zmbfbar^{(a)} \prec \zmbfbar^{(c)} \succ \zmbfbar^{(d)}$) Set $\theta_j^{(b)} = \theta_j^{(d)}$.\\
\textbf{elseif} \, ($\zmbfbar^{(c)} \prec \zmbfbar^{(d)} \succ \zmbfbar^{(e)}$) Set $\theta_j^{(a)} = \theta_j^{(c)}$,  $\theta_j^{(b)} = \theta_j^{(e)}$.\\
\textbf{elseif} \, ($\zmbfbar^{(d)} \prec \zmbfbar^{(e)} \succ \zmbfbar^{(b)}$) Set $\theta_j^{(a)} = \theta_j^{(d)}$.\\
\textbf{else} 
 Set $\theta_j^{(a)} = \theta_j^{(d)}$.\\
\textbf{Output:} $[\theta_j^{(a)}, \theta_j^{(b)}]$.  
\normalsize \vspace{0.0cm}
}}
  \end{minipage} \hspace{0.3em}
  \begin{minipage}[h]{.42\textwidth}
     \centering
\fbox{\parbox[t]{0.95\textwidth}{\vspace{0.45cm}
\begin{tikzpicture}[scale = 2.8]
    

    	\begin{scope}[shift={(-5.0,0)},scale = 0.5]\scriptsize

\def\r{0.06};
    
    \draw[thick] (0,0) .. controls (0.2,0) and (0.4,1.6) .. (0.8,1.6) 
    ..controls (1.4,1.6) and (2.2,0) .. (4,0);
    
    \draw[-latex] (0,-.1)--(0,2.505); 
    \draw[-latex] (-0.1,0)--(4.4,0);
    \node[left] at (0,2.35) {$\xi$};
    \node[below right] at (4.1,0) {$\theta_j$};
   
   	\coordinate (C1) at (0,0.00);
    \coordinate (C2) at (1,1.56);
    \coordinate (C3) at (2,0.76);
    \coordinate (C4) at (3,0.18);
    \coordinate (C5) at (4,0.00);
    
    \node[below] at (0,0) {$\theta_j^{(a)}$};
    \node[below] at (1,0) {$\theta_j^{(c)}$};
    \node[below] at (2,0) {$\theta_j^{(d)}$};
    \node[below] at (3,0) {$\theta_j^{(e)}$};
    \node[below] at (4,0) {$\theta_j^{(b)}$};
    
    \foreach \x in {1,2,3,4} {
    	\draw (\x,-.1) -- (\x,.1);
        \draw[dotted] (\x,0) -- (\x,2);
    }
    \fill[color=black] 
    		(C1) circle (\r)
    		(C2) circle (\r)
            (C3) circle (\r)
            (C4) circle (\r)
            (C5) circle (\r);   
    
    \draw[thick,-latex] (0.3,0.4) -- (0.7,0.8);
    \draw[thick,-latex] (1.7,0.4) -- (1.3,0.8);
    \draw[thick,-latex] (2.7,0.4) -- (2.3,0.8);
    \draw[thick,-latex] (3.7,0.4) -- (3.3,0.8);
    
    \draw (0,1.8)--(0,2.2) (2,1.8)--(2,2.2);
    \draw[<->, dashed, thick] (0,2)--(2,2);
    
    \end{scope}
    
\end{tikzpicture}
\vspace{0.45cm}}
}
  \end{minipage}
\end{minipage}
\caption{(Left): Subroutine \emph{ShrinkInterval}. (Right): Visual intuition of the subroutine \emph{ShrinkInterval}~\cite{hiranandani2019multiclass}; the subroutine shrinks the current interval to half based on oracle responses to the four queries.}
\label{append:fig:shrink1}
\end{figure}
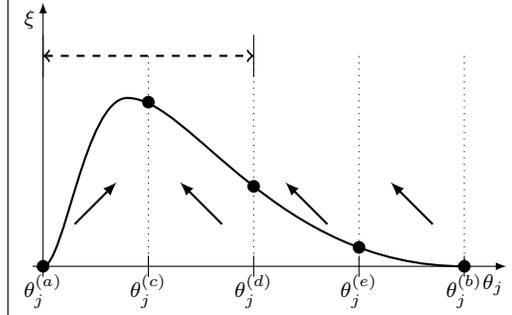

\addtocounter{algorithm}{1}
\balgorithm[t]
\caption{Linear Performance Metric Elicitation}
\label{alg:slme}
\small
\balgorithmic[1]
\STATE \textbf{Input:} Query space $\Scal$, binary-search tolerance $\epsilon > 0$, oracle $\Omega'$ with metric $\xi$\\ \hfill\\
\FOR{$i = 1, 2, \cdots q$}
\STATE Set $\ambf = \ambf' = (1/\sqrt{q}, \dots, 1/\sqrt{q})$.
\STATE Set $a'_i = -1/\sqrt{q}$.
\STATE Compute the optimal $\zbar^{(\ambf)}$ and $\zbar^{(\ambf')}$ over the sphere $\Scal$ using Lemma~\ref{lem:spherebayes}
\STATE Query $\Omega'(\zmbfbar^{(\ambf)}, \zmbfbar^{(\ambf')})$\\
\ENDFOR
\COMMENT{Fix the search orthant based on the above oracle responses}\\ \hfill \\
\STATE\textbf{Initialize:} $\bm{\theta} = \bm{\theta}^{(1)}$ \hfill \COMMENT{$\bm{\theta}^{(1)}$ is any point in the search orthant.}
\FOR{$t=1, 2, \cdots, T=4(q-1)$}
\STATE Set $\bm{\theta}^{(a)} = \bm{\theta}^{(c)}=\bm{\theta}^{(d)}=\bm{\theta}^{(e)}=\bm{\theta}^{(b)} = \bm{\theta}^{(t)}$.\\
\WHILE{$\abs{\theta^{(b)}_j - \theta^{(a)}_j} > \epsilon$}
\STATE Set $\theta^{(c)}_j = \frac{3 \theta^{(a)}_j + \theta^{(b)}_j}{4}$, $\theta^{(d)}_j = \frac{\theta^{(a)}_j + \theta^{(b)}_j}{2}$, and $\theta^{(e)}_j = \frac{\theta^{(a)}_j + 3 \theta^{(b)}_j}{4}$.
\STATE Set $\zmbfbar^{(a)} = \mu(\bm{\theta}^{(a)})$ (i.e. parametrization of $\partial \Scal$). Similarly, set $\zmbfbar^{(c)}, \zmbfbar^{(d)}, \zmbfbar^{(e)}, \zmbfbar^{(b)}$
\STATE Query $\Omega'(\zmbfbar^{(c)}, \zmbfbar^{(a)}), \Omega'(\zmbfbar^{(d)},  \zmbfbar^{(c)})$, $\Omega'(\zmbfbar^{(e)}, \zmbfbar^{(d)}),\Omega'(\zmbfbar^{(b)}, \zmbfbar^{(e)})$.
\STATE $[\theta^{(a)}_j, \theta^{(b)}_j] \leftarrow$ \emph{ShrinkInterval} (responses)\hfill \COMMENT{see Figure~\ref{append:fig:shrink1}}
\ENDWHILE
\STATE Set $\theta^{(d)}_j = \frac{1}{2}(\theta^{(a)}_j+\theta^{(b)}_j)$ \\
\STATE Set $\bm{\theta}^{(t)} = \bm{\theta}^{(d)}$.
\ENDFOR
\STATE \textbf{Output:} $\hat a_i =\Pi_{j=1}^{i-1} \sin\theta_j^{(T)} \cos{\theta_i}^{(T)} \, \forall i \in [q-1],\;\hat a_q =\Pi_{j=1}^{q-1} \sin\theta_j^{(T)}$
\ealgorithmic
\ealgorithm

\emph{Description of Algorithm~\ref{alg:slme}:}\footnote{The superscripts in Algorithm~\ref{alg:slme} denote iterates. Please do not confuse it with the sensitive group index.} Suppose that the oracle's linear metric is $\xi$ parametrized by $\ambf$ where $\Vert \ambf \Vert_2=1$ (Section~\ref{ssec:mpme}). Using the parametrization $\mu(\thetambf)$ of the surface of the sphere $\partial \Scal$ as explained above,  Algorithm~\ref{alg:slme} returns an estimate $\ambfhat$ with $ \Vert \ambfhat \Vert_2=1$. Line 2-6 in Algorithm~\ref{alg:slme} recovers the orthant of the optimal statistic over the sphere by posing $q$ trivial queries. Once the search orthant of the optimal statistic is fixed, the procedure is same as Algorithm 2 of~\cite{hiranandani2019multiclass}. In each iteration of the for loop, the algorithm updates one angle $\theta_j$ keeping other angles fixed 
by a binary-search procedure, where the \emph{ShrinkInterval}  subroutine (illustrated in Figure~\ref{append:fig:shrink1}) shrinks the interval $[\theta^a_j, \theta^b_j]$ by half based on the  responses. Then the algorithm cyclically updates each angle until it converges to a metric sufficiently close to the true metric. The number of cycles in coordinate-wise search is fixed to four.

\vspace{-0.2cm}
\section{Proofs and Details of Section~\ref{sec:confusion}}
\label{append:sec:confusion}
\vskip -0.2cm
\bproof[Proof of Proposition~\ref{prop:C}]
The set of rates $\Rcal^g$ for a group $g$ satisfies the following properties:
\bitemize[leftmargin=1em]
\item \emph{Convex}: Let us take two classifiers $h_1^g, h_2^g \in \Hcal^g$ which achieve the rates $\rmbf_1^g, \rmbf_2^g \in \Rcal^g$. We need to check whether or not  the convex combination $\alpha \rmbf_1^g + (1-\alpha)\rmbf_2^g$ is feasible, i.e., there exists some classifier which achieve this rate. Consider a classifier $h^g$, which with probability $\alpha$ predicts what classifier $h_1^g$  predicts and with probability $1-\alpha$ predicts what classifier $h_2^g$ predicts. Then the elements of the rate matrix $R_{ij}^g(h)$ is given by: 

\begin{align*}
R_{ij}^g(h) &= \Pmbb(h^g=j | Y=i)  \\ \nonumber
&= \Pmbb(h^g_1=j|h^g=h^g_1, Y=i)\Pmbb(h^g=h_1^g) + \Pmbb( h_2^g=j|h^g=h_2^g, Y=i)\Pmbb(h^g=h^g_2)   \\ \nonumber
&= \alpha \rmbf_{1}^g + (1-\alpha)\rmbf_{2}^g.
\end{align*}
Therefore, $\Rcal^g \; \forall \; g \in [m]$ is convex.
\item \emph{Bounded:} Since $R^g_{ij}(h) = P[h=j|Y=i] = P[h=j, Y=i]/P[Y=i] \leq 1$ for all $i,j \in [k]$, $\Rcal^g \subseteq [0, 1]^q$.
\item \emph{$\embf_i$'s and $\ombf$ are always achieved:} The classifier which always predicts class $i$, will achieve the rate $\embf_i$. Thus, $\embf_i \in \Rcal^g \, \forall\, i \in [k], g \in [m]$ are feasible. Just like the convexity proof, a classifier which predicts similar to one of the trivial classifiers with probability $1/k$ will achieve the rates $\ombf$. 
\item \emph{$\embf_i$'s are vertices:} Any supporting hyperplane with slope $\ell_{1i} < \ell_{1j} < 0$ and $\ell_{1p}=0$ for $p \in [k], p \neq i, j$ will be supported by $\embf_1$ (corresponding to the trivial classifier which predict class 1). Thus, $\embf_i$'s  are vertices of the convex set. As long as the class-conditional distributions are not identical, i.e., there is some signal for non-trivial classification conditioned on each group~\cite{hiranandani2019multiclass}, one can construct a ball around the trivial rate $\ombf$ and thus $\ombf$ lies in the interior.
\eitemize
\vspace{-0.4cm}
\eproof

\vspace{-1cm}
\subsection{Finding the Sphere $\Scal_\rho$}
\label{append:ssec:sphere}
\balgorithm[t]
\caption{Obtaining the sphere $\Scal_\rho$ with radius $\rho$}
\label{alg:sphere}
\small
\balgorithmic[1]
\STATE \textbf{Input:} The center $\ombf$ of the feasible region of rates across groups.
\FOR{$j=1, 2, \cdots, q$}
\STATE Let $\mathbf r_j$ be the standard basis vector for the $j$-th dimension. 
\STATE Compute the maximum $\ell_j$ such that $\ombf + \ell_j \mathbf r_j$ is feasible for all groups by solving~\eqref{eq:op1}.
\ENDFOR
\STATE Let $CONV$ be the convex hull of $\{\ombf\pm \ell_j\mathbf r_j\}_{j=1}^{q}$.
\STATE Compute the radius $s$ of the largest ball which can fit inside of $CONV$, centered at $\ombf$.
\STATE\textbf{Output:} Sphere $\Scal_\rho$ with radius $\rho=s$ centered at $\ombf$.
\ealgorithmic
\ealgorithm

In this section, we discuss how a sufficiently large sphere $\Scal_\rho$ with radius $\rho$ may be found. The following discussion is extended from~\cite{hiranandani2019multiclass} to multiple groups setting and provided here for completeness. 

The following optimization problem is a special case of OP2 in~\cite{narasimhan2018learning,Tavker+2020}. The problem corresponds to feasiblity check problem for a given rate $\rmbf_0$ achieved by all groups within small error $\epsilon >0$. 

\begin{align}
    \min_{\rmbf^g \in \Rcal^g \, \forall g \in [m]} \; 0 \qquad s.t. \;\; \Vert \rmbf^g - \rmbf_0 \Vert_2 \leq \epsilon \quad \forall \; g \in [m].
     \tag{OP1}
    \label{eq:op1}
\end{align}

The above problem checks the feasibility and if a solution to the above problem exists, then Algorithm~1 of~\cite{narasimhan2018learning} returns it. The approach in~\cite{narasimhan2018learning} constructs a classifier whose group-wise rates are $\epsilon$-close to the given rate $\rmbf_0$. 

Furthermore, Algorithm~\ref{alg:sphere} computes a value of $\rho\geq \tilde{s}/k$, where $\tilde{s}$ is the radius of the largest ball contained in the set $\Rcal^1 \cap \cdots \cap \Rcal^m$. Notice that the approach in~\cite{narasimhan2018learning} is consistent, thus we should get a good estimate of the sphere, provided we have sufficient samples. The algorithm runs offline and does not impact query complexity.

\blemma\cite{hiranandani2019multiclass}
    Let $\tilde{s}$ be the radius of the largest ball centered at $\ombf$ in $\Rcal^1 \cap \cdots \cap \Rcal^m$. Then Algorithm~\ref{alg:sphere} returns a radius $\rho\geq \tilde{s}/k$.
\elemma
\begin{proof}
    Let $\ell_j$ be as computed in the algorithm and $\ell:= \min_j \ell_j$, then we have $\ell\geq \tilde{s}$. Moreover, the region $CONV$ contains the convex hull of $\{o\pm \ell\mathbf e_j\}_{j=1}^{q}$; however, this region contains a ball of radius $\ell/\sqrt{q} = \ell/\sqrt{k^2-k}\geq \ell/k\geq \tilde{s}/k$, and thus $\rho\geq \tilde{s}/k$.
\end{proof}

\section{Derivations of Section~\ref{sec:me}}
\label{append:sec:me}

Notice that $\sum_{g=1}^m \taumbf^g = \mathbf{1}$, i.e., the vector of ones. 

\subsection{Eliciting the Misclassification Cost $\bphi(\rmbf)$; Part 1 in Figure~\ref{fig:workflow} and line 1 in Algorithm~1}
\label{append:ssec:phi}

The key to eliciting $\bphi$ is to remove the effect of fairness violation $\bvarphi$ in the oracle responses. As explained in Section~\ref{ssec:elicitphi}, we run the LPME procedure (Algorithm~\ref{alg:slme}) with the $q$-dimensional query space $\Scal_\rho$, binary search tolerance $\epsilon$, the equivalent oracle $\Omega^{\text{class}}$. From Remark~\ref{rm:ratio}, this subroutine returns a slope $\fmbf$ with $\Vert \fmbf \Vert_2=1$ such that:
\vspace{-0.2cm}
\begin{equation}
    \frac{(1-\lambdabar)a_i}{(1-\lambdabar)a_j} = \frac{f_i}{f_j} \implies \frac{a_i}{a_j} = \frac{f_i}{f_j}.
    \label{append:eq:phisolutionf}
\end{equation}
\vskip -0.2cm
Thus, we set $\ambfhat \coloneqq \fmbf$ (line~1, Algorithm~\hyperlink{alg:f-me}{1}). 

\subsection{Eliciting the Fairness Violation $\bvarphi(\tupr)$; Part 2 in Figure~\ref{fig:workflow} and lines 2-11 in Algorithm~1}
\label{append:ssec:varphi}

\subsubsection{Eliciting the Fairness Violation $\bvarphi(\tupr)$ for $m=2$; lines 2-5 in Algorithm~1}
\label{append:sssec:elicitvarphim2}

For $m=2$, we have only one vector of unfairness weights $\bmbf^{12}$, which we now aim to elicit given $\ambfhat$. As discussed in Section~\ref{ssec:elicitvarphim2}, we fix trivial rates (through trivial classifiers) to one group and allow non-trivial rates from $\Scal_\rho$ on another group. This essentially makes the metric in Definition~\ref{def:linear} linear. The elicitation procedure is as follows. 

Fix trivial classifier predicting class $1$ for group 2 i.e. fix $h^{2}(x) = 1 \, \forall \, x \in \Xcal$, and thus $\rmbf^{2}  = \embf_1$. For group 1, we constrain the confusion rates to lie in the sphere $\Scal_\rho$ i.e. $\rmbf^{1}  = \smbf$ for $\smbf \in \Scal_\rho$. 
Then the metric in Definition~\ref{def:linear} amounts to:
    \begin{align*}
    \bPsi((\smbf, \embf_1); \ambfbar, \bmbfbar^{12}, \lambdabar) &=
    (1-\lambdabar)\inner{\ambfbar \odot (1-\bm{\tau}^{2})}{\smbf} + \lambdabar  \inner{\bmbfbar^{12}}{\vert \embf_1 - \smbf \vert} + c_1.
    \numberthis \label{append:eq:metricbm2}
    \end{align*}
The above is a function of $\smbf \in \Scal_\rho$. Since $\embf_i$'s are binary vectors and since $0 \leq \smbf \leq 1$, the sign of the absolute function with respect to $\smbf$ can be recovered. Recall that the rates are defined in row major form of the rate matrices, thus $\embf_1$ is $1$ at every $(k + j*(k-1))$-th coordinate, where $j \in \{0,\dots,k-2\}$, and 0 otherwise. The coordinates where the confusion rates are $1$ in $\embf_1$, the absolute function  opens with a negative sign (wrt. $\smbf$) and with a positive sign otherwise. In particular, define a $q$-dimensional vector $\wmbf_1$ with entries $-1$ at every $(k + j*(k-1))$-th coordinate, where $j \in \{0,\dots,k-2\}$, and $1$ otherwise. One may then write the metric $\bPsi$ as:
\begin{align*}
&\bPsi((\smbf, \embf_1)\,;\, \ambfbar, \bmbfbar^{12}, \lambdabar) =  
\inner{(1-\lambdabar)\ambfbar \odot (\bm{1}-\bm{\tau}^{2}) + \lambdabar \wmbf_1 \odot 
\bmbfbar^{12}}{\smbf} + c_1.
\numberthis \label{append:eq:metricbrich1m2}
\end{align*} 
This is again a linear metric elicitation problem where $\smbf \in \Scal$. We may again use the LPME procedure (Algorithm~\ref{alg:slme}), which outputs a (normalized) slope $\breve \fmbf$ with $\Vert \breve \fmbf \Vert_2 = 1$ in line 3 of Algorithm~1. Using Remark~\ref{rm:ratio}, we get $q-1$ independent equations and may represent every element of $\bmbfbar^{12}$ based on one element, say $\bbar^{12}_{k-1}$, i.e.:
\begin{align*}
    \frac{\breve f_{k-1}}{\breve f_{i}} &= \frac{(1-\lambdabar){(1-\tau^2_{k-1})\abar_{k-1} + \lambdabar \bbar^{12}_{k-1}}}{(1-\lambdabar){(1-\tau^2_{i})\abar_{i} + \lambdabar w_{1i}\bbar^{12}_{i}}} \qquad \forall \; i \in [q].\\
    \implies 
    \lambdabar \bmbfbar^{12} &= \wmbf_1 \odot \left[ \left( \frac{(1-\lambdabar)(1-\tau^{2}_{k-1})\abar_{k-1} + \lambdabar \bbar^{12}_{k-1}}{\breve f_{k-1}}\right) \breve \fmbf - (1-\lambdabar)((1-\taumbf^{2})\odot\ambfbar)  \right].
    \numberthis \label{append:eq:metricbfirstm2}
\end{align*}

In order to elicit entire $\bmbfbar^{12}$, we need one more linear relation such as~\eqref{append:eq:metricbfirstm2}. So, we now fix the trivial classifier predicting class $k$ for group 2 i.e. fix $h^{2}(x) = k \, \forall \, \xmbf \in \Xcal$, and thus $\rmbf^{2}  = \embf_k$. For group 1, we constrain the rates to again lie in the sphere $\Scal_\rho$ i.e. $\rmbf^{1}  = \smbf$ for $\smbf \in \Scal_\rho$. 
Since the rate vectors are in row major form of the rate matrices, notice that $\embf_k$ is $1$ at every $(k-1 + j*(k-1))$-th coordinate, where $j \in \{0,\dots,k-2\}$, and 0 otherwise. 
In particular, define a $q$-dimensional vector $\wmbf_k$ with entries $-1$ at every $(k-1 + j*(k-1))$-th coordinate, where $j \in \{0,\dots,k-2\}$, and $1$ otherwise. One may then write the metric $\bPsi$ as:
\begin{align*}
&\bPsi((\smbf, \embf_k); \ambfbar, \bmbfbar^{12}, \lambdabar) =
    (1-\lambdabar)\inner{\ambfbar \odot (1-\bm{\tau}^{2})}{\smbf} + \lambdabar  \inner{\bmbfbar^{12}}{\vert \embf_k - \smbf \vert} + c_k.
\numberthis \label{append:eq:metricbrich2m2}
\end{align*}
This is a linear metric elicitation problem where $\smbf \in \Scal$. Thus, line 4 of Algorithm~1 applies LPME subroutine (Algorithm~\ref{alg:slme}), which outputs a (normalized) slope $\tilde \fmbf$ with $\Vert \tilde \fmbf \Vert_2 = 1$. Using Remark~\ref{rm:ratio}, we extract the following relation between two of its coordinates, say the $(k-1)$-th and $((k-1)^2+1)$-th coordinates:
\begin{align*}
\frac{\tilde f_{k-1}}{\tilde f_{(k-1)^2+1}} = \frac{(1-\lambdabar)(1 - \tau^{2}_{k-1})\abar_{k-1} - \lambdabar \bbar^{12}_{k-1}}{(1-\lambdabar)(1- \tau^{2}_{(k-1)^2+1}) \abar_{(k-1)^2+1} + \lambdabar \bbar^{12}_{(k-1)^2+1}}.
\numberthis \label{append:eq:metricbsecondm2}
\end{align*}
Combining equations~\eqref{append:eq:metricbfirstm2} and~\eqref{append:eq:metricbsecondm2} and replacing the true $\ambfbar$ with the estimated $\ambfhat$ from Section~\ref{ssec:elicitphi}, we have an estimate of the scaled substitute as:
\begin{align*}
    \tilde \bmbf^{12} &= \wmbf_1 \odot \left[ \delta \breve \fmbf^{12} - \ambfhat \odot (\bm{1} - \taumbf^{2})  \right], \numberthis \label{append:eq:ellsolm2}
    \\
    \text{where} \; \delta &= \frac{2(1-\tau^{2}_{k-1})\hat a_{k-1}}{\breve f_{k - 1}} \left[ \frac{ \frac{(1-\tau^{2}_{(k-1)^2+1})\hat a_{(k-1)^2+1}}{(1-\tau^{2}_{k-1})\hat a_{k-1}} -  \frac{\tilde f_{(k-1)^2+1}}{\tilde f_{k-1}} }{\left( \frac{\breve f_{(k-1)^2+1}}{\breve f_{k-1}} - \frac{\tilde f_{(k-1)^2+1}}{\tilde f_{k-1}} \right)} \right]
\end{align*}
and $\tilde \bmbf$ is a scaled substitute defined as $\tilde \bmbf^{12}\coloneqq \frac{\lambdabar}{(1-\lambdabar)} \bmbfbar^{12}$, which nonetheless is computable from~\eqref{append:eq:ellsolm2}. Since we require a solution $\bmbfhat$ such that $\Vert \bmbfhat \Vert_2 = 1$ (Definition~\ref{def:linear}), we normalize $\tilde \bmbf$ and get the final solution:
\begin{equation}
\bmbfhat^{12} = \frac{\tilde \bmbf^{12}}{\Vert \tilde \bmbf^{12} \Vert_2}.
 \label{append:eq:bsolm2}
\end{equation} 
Notice that, due to the above normalization, the solution is  independent of the true trade-off $\lambdabar$.

\subsubsection{Eliciting the Fairness Violation $\bvarphi(\tupr)$ for $m>2$; line 6-11 in Algorithm~1}
\label{append:sssec:elicitvarphi}

Let $\Mcal \subset 2^{[m]} \setminus \{\varnothing, [m]\}$ be a set of subsets of the $m$ groups such that each element $\sigma \in \Mcal$ and $[m] \setminus \sigma$  partition the set of $m$ groups. For example, when the number of groups $m=3$, we may choose $\Mcal = \{\{1,2\}, \{1,3\}, \{2,3\}\}$. 
We will later discuss how to choose $\Mcal$ for efficient elicitation.
When $m>2$, we partition the set of groups $[m]$ into two sets of groups. Let $\sigma \in \Mcal$ and $[m] \setminus \sigma$ be one such partition of the $m$ groups defined by the set $\sigma$. We follow exactly similar procedure as in the previous section i.e. fixing trivial rates (through trivial classifiers) on the groups in $\sigma$ and allowing non-trivial rates from $\Scal_\rho$ on the groups in $[m] \setminus \sigma$. In particular, consider a paramterization $\nu : (\Scal_\rho, \Mcal, [k]) \rightarrow \Rcal^{1:m}$ defined as:
\begin{equation}
    \nu(\smbf, \sigma, i) \coloneqq \tupr \quad \text{such that} \quad \rmbf^g = \begin{cases}
    \embf_i & \text{if } g \in \sigma\\
    \smbf & \text{o.w. }
    \end{cases}
\label{append:eq:parvarphi}
\end{equation}
i.e., $\nu$ assigns trivial confusion rates $\embf_i$ on the groups in $\sigma$ and assigns $\smbf \in \Scal_\rho$ on the rest of the groups. 
Similar to the previous section, we first fix trivial classifier predicting class $1$ for groups in $\sigma$ and constrain the rates for groups in $[m] \setminus \sigma$ to be on the sphere $\Scal_\rho$. Such a setup is governed by the parametrization $\nu(\cdot,\sigma, 1)$ in equation~\eqref{append:eq:parvarphi}. Specifically, fixing $h^g(\xmbf)=1\; \forall\; g \in \sigma$ would entail the metric in Definition~\ref{def:linear} to be:
    \begin{align*}
    \bPsi(\nu(\smbf, \sigma, 1); \ambfbar, \Bmbfbar, \lambdabar) &= 
    (1-\lambdabar)\inner{\ambfbar\odot(\bm{1} - \taumbf^{\sigma})}{\smbf} +  \lambda \inner{\etambfbar^{\sigma}}{\vert \embf_1 - \smbf \vert} + c_1,
    \numberthis \label{append:eq:metricb}
    \end{align*}
where $\taumbf^{\sigma} = \sum_{g\in \sigma}\bm{\tau}^{g}$ and $\etambfbar^{\sigma} = \sum_{u, v \in [m], v > u} \1\left[|\{u,v\}\cap\sigma|=1\right]\bmbfbar^{uv}$. 
Similar to the previous section, since $\embf_i$'s are binary vectors, the sign of the absolute function wrt. $\smbf$ can be recovered. In particular, the metric amounts to:
\begin{align*}
&\bPsi(\nu(\smbf, \sigma, 1); \ambfbar, \Bmbfbar, \lambdabar) = \inner{(1-\lambdabar)\ambfbar \odot (\bm{1}-\bm{\tau}^{2}) + \lambdabar \wmbf_1 \odot 
\etambfbar^{\sigma}}{\smbf} + c_1,
\numberthis \label{append:eq:metricbrich1}
\end{align*}
where $\wmbf_1 \coloneqq 1 - 2\embf_1$ and $c_1$ is a constant not affecting the responses. Notice that~\eqref{append:eq:metricb} and~\eqref{append:eq:metricbrich1} are analogous to~\eqref{append:eq:metricbm2} and~\eqref{append:eq:metricbrich1m2}, respectively, except that $\taumbf^2$ is replaced by $\taumbf^\sigma$ and $\bmbfbar^{12}$ is replaced by $\etambfbar^{\sigma}$. This is a linear metric in $\smbf$. We again the use the LPME procedure in line 8 of Algorithm~1, which outputs a normalized slope $\breve \fmbf^\sigma$ such that $\Vert \breve \fmbf^\sigma\Vert_2=1$, and thus we get an analogous solution to~\eqref{append:eq:metricbfirstm2} as:
\begin{align*}
    \lambdabar \etambfbar^{\sigma} &= \wmbf_1 \odot \left[ \left( \frac{(1-\lambdabar)(1 - \tau^{\sigma}_{k-1})\abar_{k-1} + \lambdabar \etabar^{\sigma}_{k-1}}{\breve f^{\sigma}_{k-1}}\right) \breve \fmbf^{\sigma} - (1-\lambdabar)((\bm{1} - \taumbf^{\sigma})\odot\ambfbar  \right].
    \numberthis \label{append:eq:metricbfirst}
\end{align*}
In order to elicit entire $\etambfbar^{\sigma}$, we need one more linear relation such as~\eqref{append:eq:metricbfirst}. So, we now fix the trivial rates through trivial classifier predicting class $k$ for the groups in $\sigma$ i.e. fix $h^{g}(x) = k \, \forall \, \xmbf \in \Xcal$ if $g \in \sigma$, and thus $\rmbf^{g}  = \embf_k$ for all groups $g \in \sigma$. For the rest of the groups, we constrain the confusion rates to again lie in the sphere $\Scal_\rho$ i.e. $\rmbf^{g}  = \smbf$ for $\smbf \in \Scal_\rho$ for all groups $g \in [m] \setminus \sigma$. Such a setup is governed by the parametrization $\nu(\cdot,\sigma, k)$~\eqref{append:eq:parvarphi}. The metric $\bPsi$ in Definition~\ref{def:linear} amounts to:
\begin{align*}
&\bPsi(\nu(\smbf, \sigma, k); \ambfbar, \Bmbfbar, \lambdabar) = (1-\lambdabar)\inner{\ambfbar \odot (1-\bm{\tau}^{\sigma})}{\smbf} + \lambdabar  \inner{\etambfbar^{\sigma}}{\vert \embf_k - \smbf \vert} + c_k.
\numberthis \label{append:eq:metricbrich2}
\end{align*}
Thus by running LPME procedure again in line 9 of Algorithm~1 results in $\tilde \fmbf^{12}$ with $\Vert \tilde \fmbf^{12} \Vert_2 = 1$. Using Remark~\ref{rm:ratio}, we extract the following relation between the $(k-1)$-th and $((k-1)^2+1)$-th coordinates:
\begin{align*}
\frac{\tilde f^{\sigma}_{k-1}}{\tilde f^{\sigma}_{(k-1)^2+1}} = \frac{(1-\lambdabar)(1- \tau^{\sigma}_{k-1})\abar_ {k-1} - \lambdabar \etabar^{\sigma}_{k-1}}{(1-\lambdabar)(1 - \tau^{\sigma}_{(k-1)^2+1})\abar_{(k-1)^2+1} + \lambdabar \etabar^{\sigma}_{(k-1)^2+1}}.
\numberthis \label{append:eq:metricbsecond}
\end{align*}
Combining equations~\eqref{append:eq:metricbfirst} and~\eqref{append:eq:metricbsecond}, we have:

\begin{align*}
    \sum\nolimits_{u, v} \1\left[|\{u,v\}\cap \sigma|=1\right] \tilde \bmbf^{uv} &= \gammambf^{\sigma}, \quad \text{where} \numberthis \label{append:eq:ellsol}\\
    \gammambf^{\sigma} = \wmbf_1 \odot \left[ \delta^{\sigma} \fmbf^{\sigma} - \ambfhat \odot (\bm{1} - \taumbf^{\sigma})  \right],
    \; \delta^{\sigma} &= \frac{2(1-\tau^{\sigma}_{k-1})\hat a_{k-1}}{f^\sigma_{k - 1}} \left[ \frac{ \frac{(1 - \tau^{\sigma}_{(k-1)^2+1})\hat a_{(k-1)^2+1}}{(1-\tau^{\sigma}_{k-1})\hat a_{k-1}} -  \frac{\tilde f^{\sigma}_{(k-1)^2+1}}{\tilde f^{\sigma}_{k-1}} }{\left( \frac{f^{\sigma}_{(k-1)^2+1}}{f^{\sigma}_{k-1}} - \frac{\tilde f^{\sigma}_{(k-1)^2+1}}{\tilde f^{\sigma}_{k-1}} \right)} \right],
\end{align*}
and $\tilde \bmbf^{uv} \coloneqq \lambdabar\bmbfbar^{uv}/(1 - \lambdabar)$ is a scaled version of the true (unknown) $\bmbfbar$, which nonetheless can be computed from~\eqref{append:eq:ellsol}.

By two runs of LPME algorithm, we can get $\gammambf^{\sigma}$ and solve~\eqref{append:eq:ellsol}. However, the left hand side of~\eqref{append:eq:ellsol} does not allow us to recover the $\tilde \bmbf$'s separately and provides only one equation. Let us denote the Equation~\eqref{append:eq:ellsol} by $\ell^\sigma$ corresponding to the set $\sigma$. In order to elicit all $\tilde \bmbf$'s we need a system of $M \coloneqq {m\choose 2}$ independent equations.   
This is easily achievable by choosing $M$ $\sigma$'s so that we get $M$ set of unique equations like~\eqref{append:eq:ellsol}. Let $\Mcal$ be those set of sets. In most cases,  pairing two groups to have trivial rates (through trivial classifiers) and rest of the groups to have rates from the sphere $\Scal$ will work. For example, when $m=3$, fixing $\Mcal = \{ \{1,2\}, \{1,3\}, \{2,3\}\}$ suffices. Thus, running over all the choices of sets of groups $\sigma \in \Mcal$ provides the system of equations $\Lcal \coloneqq \cup_{\sigma \in \Mcal} \ell^\sigma$ (line 10 in Algorithm~\hyperlink{alg:f-me}{1}), which is formally described as follows:
\begin{equation}
    \left[ \begin{array}{cccc} \Xi & 0 & \dots & 0\\
    0 & \Xi & \dots & 0 \\
    \dots & \dots & \dots & \dots \\
    0 & 0 & \dots & \Xi 
    \end{array}\right] \left[ \begin{array}{c} \tilde \bmbf_{(1)} \\
    \tilde \bmbf_{(2)} \\
    \dots \\
    \tilde \bmbf_{(q)}
    \end{array}\right] = \left[ \begin{array}{c} \bm\gamma_{(1)} \\
    \bm\gamma_{(2)} \\
    \dots \\
    \bm\gamma_{(q)}
    \end{array}\right],
    \label{append:btilde}
\end{equation}
where $\tilde \bmbf_{(i)} = (\tilde b_{i}^1,\tilde b_{i}^2, \cdots, \tilde b_{i}^M)$ and $\gammambf_{(i)} = (\gamma_{i}^1, \gamma_{i}^2, \cdots, \gamma_{i}^M)$ are vectorized versions of the $i$-th entry across groups for $i \in [q]$, and $\Xi \in \{0,1\}^{M\times M}$ is a binary full-rank matrix denoting membership of groups in the set $\sigma \in \Mcal$. For instance, for the choice of $\Mcal = \{ \{1,2\}, \{1,3\}, \{2,3\}\}$ when $m=3$ gives:
$$\Xi = \left[ \begin{array}{ccc} 0 & 1 & 1\\
    1 & 0 & 1\\
    1 & 1 & 0\\
\end{array}\right].$$
From technical point of view, one may choose any $\Mcal$ such that the resulting group membership matrix $\Xi$ is non-singular. Hence the solution of the system of equations $\Lcal$ is:
\begin{equation}
    \left[ \begin{array}{c} \tilde \bmbf_{(1)} \\
    \tilde \bmbf_{(2)} \\
    \dots \\
    \tilde \bmbf_{(q)}
    \end{array}\right] = \left[ \begin{array}{cccc} \Xi & 0 & \dots & 0\\
    0 & \Xi & \dots & 0 \\
    \dots & \dots & \dots & \dots \\
    0 & 0 & \dots & \Xi
    \end{array}\right]^{(-1)} \left[ \begin{array}{c} \bm\gamma_{(1)} \\
    \bm\gamma_{(2)} \\
    \dots \\
    \bm\gamma_{(q)}
    \end{array}\right].
    \label{append:eq:sol-b}
\end{equation}
When we normalize $\tilde \bmbf$, we get the final fairness violation weight estimates as:
\begin{equation}
\bmbfhat^{uv} = \frac{\tilde \bmbf^{uv}}{\sum_{u,v=1, v > u}^m \Vert \tilde \bmbf^{uv} \Vert_2} \quad \text{for} \quad u,v \in [m], v>u.
 \label{append:eq:bsol}
\end{equation} 
Notice that, due to the above normalization, the solution is again independent of the true trade-off $\lambdabar$.

\subsection{Eliciting Trade-off $\lambdabar$; Part 3 in Figure~\ref{fig:workflow} and line 12 in Algorithm~1}
\label{append:ssec:lambda}

For ease of notation, let us construct a parametrization $\nu' : \Scal^+_\varrho  \rightarrow \Rcal^{1:m}$:
\vspace{-0.1cm}
\begin{equation}
\nu'(\smbf^+) \coloneqq (\smbf^+, \ombf, \dots, \ombf),
    \label{append:eq:parlambda}
\end{equation}

Using the parametrization $\nu'$ from~\eqref{append:eq:parlambda}, the metric in Definition~\ref{def:linear} reduces to a linear metric in $\smbf^+$ as discussed in~\eqref{eq:metriclambda}, i.e:
\begin{align*}
        \bPsi(\nu'(\smbf^+) \,;\,\ambfbar, \Bmbfbar, \lambdabar) = \inner{(1-\lambdabar)\taumbf^1\odot\ambfbar + \lambdabar \sum\nolimits_{v=2}^m \bmbfbar^{1v}}{\smbf^+} + c.
         \numberthis \label{append:eq:metriclambda}
\end{align*}

We first show the proof of Lemma~\ref{lm:lambda} and then discuss the trade-off elicitation algorithm (Algorithm~\ref{alg:lambda}).

\bproof[Proof of Lemma~\ref{lm:lambda}]

For simplicity, let us abuse notation for this proof and denote $\taumbf^1\odot\ambfbar$ simply by $\ambf$,  $\sum\nolimits_{v=2}^m \bmbfbar^{1v}$ simply by $\bmbf$, and $\Scal_\varrho^+$ simply by $\Scal$. 

$\Scal$ is a convex set. Let $\Zcal = \{\zmbf = (z_1, z_2) \,|\, z_1 = <\ambf, \smbf>, z_2 = <\bmbf, \smbf>, \smbf \in \Scal\}$.

\emph{Claim:} $\Zcal$ is convex.

Let $z, z' \in \Zcal$.

$\alpha z_1+ (1-\alpha) z'_1 ~=~
\alpha <\ambf, \smbf> + (1-\alpha)  <\ambf, \smbf'> 
~=~
<\ambf, \alpha \smbf + (1-\alpha)  \smbf'> 
$

$\alpha z_2+ (1-\alpha) z'_2 ~=~
\alpha <\bmbf, \smbf> + (1-\alpha)  <\bmbf, \smbf'> 
~=~
<\bmbf, \alpha \smbf + (1-\alpha)  \smbf'> 
$

Since $\alpha \smbf + (1-\alpha)  \smbf' \in \Scal$,
$\alpha z + (1-\alpha) z' \in \Zcal$. Hence $\Zcal$ is convex. 

\emph{Claim:} The boundary of the set $\Zcal$ is a strictly convex curve with no vertices for $\ambf\neq \bmbf$.  

Recall that, the required function is given by:
\begin{align}
\vartheta(\lambda)  = \max\nolimits_{\zmbf \in \Zcal} (1-\lambda)z_1 + \lambda z_2 + c \label{eq:metricwomod}
\end{align}

(i) Since the set $\Zcal$ is convex, every boundary point is supported by a hyperplane. 

(ii)  Since $\ambf \neq \bmbf$, notice that the slope is uniquely defined by $\lambda$. Since the sphere $\Scal$ is strictly convex, the above linear functional defined by $\lambda$ is maximized by a  unique point in $\Zcal$ (similar to Lemma~\ref{lem:spherebayes}). Thus, the the hyperplane is tangent at a unique point on the boundary of $\Zcal$. 

(iii) It only remains to show that there are no vertices on the boundary of $\Zcal$. Recall that a vertex exists if (and only if) some point is supported by more than one tangent hyperplane in two dimensional space. This means there are two values of $\lambda$ that achieve the same maximizer. This is contradictory since there are no two linear functionals that achieve the same maximizer on $\Scal$. 

This implies that the boundary of $\Zcal$ is strictly convex curve with no vertices. Since we are interested in the maximization of $\vartheta$, let us call this boundary as the upper boundary and denote it by $\partial\Zcal_+$. 

\emph{Claim:} Let  $\upsilon:[0,1]\to \partial\mathcal \Zcal_+$ be continuous, bijective, parametrizations of the upper boundary. Let $\vartheta:\mathcal \Zcal \to \mathbb R$ be a quasiconcave function which is 
monotone increasing in both $z_1$ and $z_2$.
Then the composition $\vartheta\circ \upsilon: [0,1]\to\mathbb \Rmbb$ is strictly
quasiconcave (and therefore unimodal with no flat regions) on the interval $[0, 1]$.

Let $S$ be some superlevel set of the quasiconcave function $\vartheta$. 
Since $\upsilon$ is a continuous bijection and since the boundary $\partial \Zcal_+$ is a strictly convex curve with no vertices, wlog., for any $r<s<t$, 
$z_1(\upsilon(r))< z_1(\upsilon(s)) < z_1(\upsilon(t))$, and $z_2(\upsilon(r))>z_2(\upsilon(s))>z_2(\upsilon(t))$.
(otherwise, swap $r$ and $t$). Since the boundary $\partial \Zcal_+$ is a strictly convex curve, then $\upsilon (s)$ must be greater (component-wise) a point in the convex combination of $\upsilon(r)$ and $\upsilon(t)$. Let us denote that point by $u$. Since $\vartheta$ is monotone increasing, then $x\in S$ implies that $y \in S$, too,  for all $y\geq x$ componentwise. Therefore, $\vartheta(\upsilon(s)) \leq \vartheta(u)$. Since $S$ is convex, $u \in S$ and thus $\upsilon(s) \in S$.

This implies that $\upsilon^{-1}(\partial \Zcal_+\cap S)$ is an interval; hence it is convex, which in turn tells us that the superlevel sets of $\vartheta\circ \upsilon$ are convex. So, $\vartheta\circ \upsilon$ is quasiconcave, as desired. This implies unimodaltiy, because a function  defined on real line which has more than one local maximum can not be quasiconcave. Moreover, since there are no vertices on the boundary $\partial \Zcal_+$, the $\vartheta\circ \upsilon: [0,1]\to\mathbb \Rmbb$ is strictly quasiconcave (and thus unimodal with no flat regions) on the interval $[0, 1]$. This completes the proof of Lemma~\ref{lm:lambda}.
\eproof

\balgorithm[t]
\caption{Eliciting the trade-off $\lambdabar$}
\label{alg:lambda}
\small
\balgorithmic[1]
\STATE \textbf{Input:} Query space $\Scal_\varrho^+$, binary-search tolerance $\epsilon > 0$, oracle $\Omega^{\text{trade-off}}$
\STATE \textbf{Initialize:} $\lambda^{(a)} = 0$, $\lambda^{(b)} = 1$.
\WHILE{$\abs{\lambda^{(b)} - \lambda^{(a)}} > \epsilon$} 
\STATE Set $\lambda^{(c)} = \frac{3 \lambda^{(a)} + \lambda^{(b)}}{4}$, $\lambda^{(d)} = \frac{\lambda^{(a)} + \lambda^{(b)}}{2}$, $\lambda^{(e)} = \frac{\lambda^{(a)} + 3 \lambda^{(b)}}{4}$
\STATE Set $\smbf^{(a)} = \displaystyle\argmax_{\smbf^+\in\Scal_\varrho^+} \inner{(1-\lambda_a)\taumbf^1\odot\ambfhat + \lambda_a \sum_{v=2}^m \bmbfhat^{1v}}{\smbf^+}$ using Lemma~\ref{lem:spherebayes}
\STATE Similarly, set $\smbf^{(c)}$, $\smbf^{(d)}$, $\smbf^{(e)}$, $\smbf^{(b)}$.
\STATE Query  $\Omega^{\text{trade-off}}(\smbf^{(c)}, \smbf^{(a)})$,  $\Omega^{\text{trade-off}}(\smbf^{(d)}, \smbf^{(c)})$,  $\Omega^{\text{trade-off}}(\smbf^{(e)}, \smbf^{(d)})$, and  $\Omega^{\text{trade-off}}(\smbf^{(b)}, \smbf^{(e)})$.\\
\STATE $[\lambda^{(a)}, \lambda^{(b)}] \leftarrow$ \emph{ShrinkInterval} (responses) using a subroutine analogous to the routine shown in  Figure~\ref{append:fig:shrink1}.
\ENDWHILE
\STATE \textbf{Output:} $\hat\lambda = \frac{\lambda^{(a)}+\lambda^{(b)}}{2}$. 
\ealgorithmic
\ealgorithm
\vspace{-0.3cm}
\emph{Description of Algorithm~\ref{alg:lambda}:}\footnote{The superscripts in Algorithm~\ref{alg:slme} denote iterates. Please do not confuse it with the sensitive group index.} Given the unimodality of $\vartheta(\lambda)$ from Lemma~\ref{lm:lambda}, we devise the binary-search procedure Algorithm~\ref{alg:lambda} for eliciting the true trade-off $\lambdabar$. The algorithm takes in input the query space $\Scal_\varrho^+$, binary-search tolerance $\epsilon$, an equivalent oracle  $\Omega^{\text{trade-off}}$, the elicited $\ambfhat$ from Section~\ref{ssec:elicitphi}, and the elicited $\Bmbfhat$ from Section~\ref{ssec:elicitvarphi}. The algorithm finds the maximizer of the function $\hat{\vartheta}(\lambda)$ defined analogously to~\eqref{eq:vartheta}, where $\ambfbar, \Bmbfbar$ are replaced by $\ambfhat, \Bmbfhat$. The algorithm poses four queries to the oracle and shrink the interval $[\lambda^{(a)}, \lambda^{(b)}]$ into half based on the responses using a subroutine analogous to \emph{ShrinkInterval} shown in Figure~\ref{append:fig:shrink1}. The algorithm stops when the length of the search interval $[\lambda^{(a)}, \lambda^{(b)}]$ is less than the tolerance $\epsilon$.
\vspace{-0.3cm}
\section{Proof of Section~\ref{sec:guarantees}}
\label{append:sec:guarantees}
\vskip -0.5cm
\bproof[Proof of Theorem~\ref{thm:error}] Let $\Vert \cdot \Vert_\infty$ denote the $\ell$-infinity norm. We break this proof into three parts. 
\vspace{-0.2cm}
\begin{enumerate}[leftmargin=*]
\item \emph{Elicitation guarantees for the misclassification cost $\hphi$ (i.e., $\ambfhat$)}

Since Algorithm~\hyperlink{alg:f-me}{1} elicits a linear metric using the $q$-dimensional sphere $\Scal$, the guarantees on $\ambfhat$ follows from Theorem 2 of~\cite{hiranandani2019multiclass}. Thus, under Assumption~\ref{as:regularity}, the output $\ambfhat$ from line 1 of Algorithm~\hyperlink{alg:f-me}{1} satisfies $\Vert \ambf^*-\ambfhat \Vert_{2}\leq O(\sqrt{q}(\epsilon+\sqrt{\epsilon_\Omega/\rho}))$ after $O\left(q\log \tfrac \pi {2\epsilon}\right)$ queries. 

\item \emph{Elicitation guarantees for the fairness violation cost $\hvarphi$ (i.e., $\Bmbfhat$)}

We start with the definition of true $\bm{\gamma}$ (i.e. when all the elicited entities are true) from~\eqref{append:eq:ellsol} and let us drop the superscript $\sigma$ for simplicity. Furthermore, let $\epsilon+\sqrt{\epsilon_\Omega/\rho}$ be denoted by $\epsilon$.

\begin{align*}
\gammambf = \wmbf_1 \odot \left[ \delta  \breve \fmbf - \ambfbar\odot(\bm{1}-\taumbf)  \right] \quad \text{where} \; \delta = \frac{2(1-\tau_{k-1})\abar_{k-1}}{ \breve f_{k - 1}} \left[ \frac{ \frac{(1-\tau_{(k-1)^2+1}) \abar_{(k-1)^2+1}}{(1-\tau_{k-1}) \abar_{k-1}} -  \frac{ \tilde f_{(k-1)^2+1}}{\tilde f_{k-1}} }{\left( \frac{\breve f_{(k-1)^2+1}}{ \breve f_{k-1}} - \frac{ \tilde f_{(k-1)^2+1}}{ \tilde f_{k-1}} \right)} \right].
\end{align*}
Let us look at the derivative of the $i$-th coordinate of $\bm{\gamma}$.  
$$
\frac{\partial \gamma_i}{\partial a_j} = \begin{cases}
0 & \text{if } j\neq i,j\neq k-1,j\neq(k-1)^2+1\\
-\tau_i & \text{if } j = i\\
c_{i, 1} & \text{if } j = k-1\\
c_{i, 2} & \text{if } j = (k-1)^2+1,
\end{cases}
$$
where $c_{i, 1}$ and $c_{i, 2}$ are some bounded constants due to Assumption~\ref{as:regularity}. Similarly, $\partial\gamma_i/\partial f_j$ is bounded as well due to the regularity Assumption~\ref{as:regularity}. This means that $\gamma_i$ is Lipschitz in 2-norm wrt. $\ambf$ and $\fmbf$. Thus,
$$
\Vert \bm{\gamma} - \bm{\hat \gamma} \Vert_\infty \leq c_3 \Vert \ambfbar - \ambfhat  \Vert_2  + c_4 \Vert \breve \fmbf - \hat{\breve{\fmbf}} \Vert_2,
$$
for some Lipschits constants $c_3$ and $c_4$. From the bounds of Part~1 of this proof, we have:
$$
\Vert \bm{\gamma} - \bm{\hat \gamma} \Vert_\infty \leq O(\sqrt q\epsilon).
$$

Recall the construction of $\tilde \bmbf_{(i)}$ from~\eqref{append:btilde}. We then have from the solution of system of equations~\eqref{append:eq:sol-b} that:

$$
 \tilde \bmbf_{(i)} = \Xi^{-1}\bm{\gamma}_{(i)} \quad \forall \; i \in [q],
$$
where $\tilde \bmbf_{(i)} =  (\tilde b_{i}^1, \tilde b_{i}^2, \cdots, \tilde b_{i}^M)$ and $\tilde \gammambf_{(i)} = (\gamma_{i}^1, \gamma_{i}^2, \cdots, \gamma_{i}^M)$ are vectorized versions of the $i$-th entry across groups for $i \in [q]$. $\Xi \in \{0,1\}^{M \times M}$ is a full-rank symmetric matrix with bounded infinity norm $\Vert \Xi^{-1} \Vert_\infty \leq c$ (here, infinity norm of a matrix is defined as the maximum absolute row sum of the matrix). Thus we have:

$$
\Vert \tilde \bmbf_{(i)} - \hat {\tilde\bmbf}_{(i)}\Vert_\infty = \Vert \Xi^{-1} \bm{\gamma}_{(i)} - \Xi^{-1} \hat{\bm{\gamma}}_{(i)}
\Vert_\infty = \Vert \Xi^{-1} (\bm{\gamma}_{(i)} - \hat{\bm{\gamma}}_{(i)}) \Vert_\infty \leq \Vert \Xi^{-1} \Vert_\infty \Vert \bm{\gamma}_{(i)} - \hat{\bm{\gamma}}_{(i)} \Vert_\infty,
$$
which gives
$$
\Vert \tilde \bmbf_{(i)} - \hat {\tilde\bmbf}_{(i)}\Vert_\infty \leq O(\sqrt q \epsilon).
$$

Now, our final estimate is the normalized form of $\hat{\tilde{\bmbf}}$ from~\eqref{append:eq:bsol}, so the final error in the stacked version $vec(\Bmbfbar)$ and $vec(\Bmbfhat)$ is:






\begin{equation}
\Vert vec(\Bmbfbar) - vec(\Bmbfhat) \Vert_\infty \leq O(\sqrt q\epsilon).
\label{eq:errinB}
\end{equation}

Since there are $q\times M$ entities in $vec(\Bmbf)$, we have:

\begin{align*}
\Vert vec(\Bmbfbar) - vec(\Bmbfhat) \Vert_2 \leq O(\sqrt{qM}\sqrt q\epsilon) = O(mq\epsilon).
\numberthis \label{append:bguarantee}
\end{align*}

Due to elicitation on sphere and the oracle noise $\epsilon_\Omega$ as defined in Definition~\ref{def:noise}, we can replace $\epsilon$ with $\epsilon + \sqrt{\epsilon_\Omega/\rho}$ back to get the final bound on fairness violation weights as in Theorem~\ref{thm:error}.

\item \emph{Elicitation guarantees for the trade-off parameter (i.e., $\lambdahat$)}

The metric for our purpose is a linear metric in $\smbf^+ \in \Scal_\rho^+$ with the following slope:
\vspace{-0.2cm}
\begin{align*}
    \bPsi(\nu'''(\smbf^+) \,;\,\ambfbar, \Bmbfbar, \lambdabar) = \inner{(1-\lambdabar)\taumbf^{1}\odot\ambfbar + \lambdabar \sum_{v=2}^m \bmbfbar^{1v}}{\smbf^+}. \numberthis
    \label{append:eq:lambdametric}
\end{align*}
\vskip -0.2cm
Since we elicit $\lambda$ through queries over a surface of the sphere, we pose this problem as finding the right angle (slope) defined by the true $\lambdabar$. Note that
$\lambdabar$ is what we want to elicit; however, due to oracle noise $\epsilon_\Omega$, we can only aim to achieve a target angle $\lambda_t$. Moreover, we do not have true $\ambfbar$ and $\Bmbfbar$ but have only estimates $\ambfhat$ and $\Bmbfhat$. Thus we query proxy solutions always and can only aim to achieve an estimated version $\lambda_e$ of the target angle. Lastly, Algorithm~\ref{alg:lambda} is stopped within an $\epsilon$ threhsold, thus the final solution $\lambdahat$  is within $\epsilon$ distance from $\lambda_e$. In total, we want to find:

$$
\vert \lambdabar - \lambdahat \vert \leq \underbrace{\vert \lambdabar - \lambda_t \vert}_{\text{oracle error}} +  \underbrace{\vert \lambda_t - \lambda_e \vert}_{\text{estimation error}} + \underbrace{\vert \lambda_e - \lambdahat \vert}_{\text{optimization error}}.
$$

\begin{itemize}[itemsep = 0pt, leftmargin = 0.5cm]
    \item optimization error: $\vert \lambda_e - \lambdahat \vert\leq \epsilon$. 
    \item oracle error: Notice that the oracle correctly answers as long as $\varrho(1 - \cos(\lambdabar - \lambda_t)) > \epsilon_\Omega$. This is due to the fact that the metric is a 1-Lipschitz linear function, and the optimal value on the sphere of radius $\varrho$ is $\varrho$. However, as $1 - \cos(x) \geq x^2/3$, so oracle is correct as long as $\vert \lambdabar - \lambda_e\vert \geq \sqrt{3\epsilon_\Omega/\varrho}$. Given this condition, the binary search proceeds in the correct direction. 
    \item estimation error: We make this error because we only have access to the estimated $\ambfhat$ and $\Bmbfhat$ not the true $\ambfbar$ and $\Bmbfbar$. However, since the metric in~\eqref{append:eq:lambdametric} is Lipschitz in $\ambfbar$ and $\sum_{v=2}^m \bmbfbar^{1v}$, this error can be treated as oracle feedback noise where the oracle responses with the estimated $\ambfhat$ and $\Bmbfhat$. Thus, if we replace $\epsilon_\Omega$ from the previous point to the error in $\ambfhat$ and $\sum_{v=2}^m\bmbfhat^{1v}$, the binary search Algorithm~\ref{alg:lambda} moves in the right direction as long as 
    $$
    \vert \lambda_t -\lambda_e\vert \geq O\left(\sqrt{\frac{\Vert \ambfbar - \ambfhat \Vert_2 + \sum_{v=2}^m \Vert \bmbfbar^{1v} - \bmbfhat^{1v} \Vert_2}{\varrho}}\right) = O\left(\sqrt{m q (\epsilon + \sqrt{\epsilon_\Omega/\rho})/\varrho}\right),
    $$
    where we have used~\eqref{append:bguarantee} to bound the error in $\{\bmbfhat^{1v}\}_{v=2}^m$.
\end{itemize}
Combining the three error bounds above gives us the desired result for trade-off parameter in Theorem~\ref{thm:error}.
\end{enumerate}
\vspace{-0.4cm}
\eproof

\end{appendices}